\documentclass{article}

\usepackage[final]{corl_2020} 

\usepackage{times}
\usepackage{graphicx}
\usepackage{wrapfig, lipsum, booktabs}
\usepackage{xcolor}
\usepackage{amsmath}
\usepackage{amsfonts}
\usepackage{caption}
\usepackage{subcaption}
\usepackage{algorithm2e}
\usepackage{algorithm}
\usepackage{algorithmic}
\usepackage[bottom]{footmisc} 
\usepackage{color}
\usepackage{setspace}

\newcommand{\task}{\mathcal{T}}

\newcommand{\E}{\mathbb{E}}

\newcommand{\z}{\mathbf{z}}
\newcommand{\obs}{\mathbf{x}}
\newcommand{\state}{\mathbf{z}}
\newcommand{\tstate}{\mathbf{s}}
\newcommand{\act}{\mathbf{a}}

\newcommand{\Method}{MELD } 
\newcommand{\MethodNoSpace}{MELD}
\newcommand{\be}{\mathbf{b}}

\newcommand\blfootnote[1]{%
  \begingroup
  \renewcommand\thefootnote{}\footnote{#1}%
  \addtocounter{footnote}{-1}%
  \endgroup
}

\usepackage[compact]{titlesec} 
\titlespacing{\section}{0pt}{*2}{*0}
\title{MELD: Meta-Reinforcement Learning from Images via Latent State Models}

\author{
  Tony Z. Zhao\textsuperscript{* \textdagger}, Anusha Nagabandi\textsuperscript{* \textdagger}, Kate Rakelly\textsuperscript{* \textdagger}, Chelsea Finn\textsuperscript{\textdaggerdbl}, Sergey Levine\textsuperscript{\textdagger} 
}

\begin{document}
\maketitle


\begin{abstract}
Meta-reinforcement learning algorithms can enable autonomous agents, such as robots, to quickly acquire new behaviors by leveraging prior experience in a set of related training tasks. However, the onerous data requirements of meta-training compounded with the challenge of learning from sensory inputs such as images have made meta-RL challenging to apply to real robotic systems. Latent state models, which learn compact state representations from a sequence of observations, can accelerate representation learning from visual inputs. In this paper, we leverage the perspective of meta-learning as task inference to show that latent state models can \emph{also} perform meta-learning given an appropriately defined observation space. Building on this insight, we develop meta-RL with latent dynamics (MELD), an algorithm for meta-RL from images that performs inference in a latent state model to quickly acquire new skills given observations and rewards. MELD outperforms prior meta-RL methods on several simulated image-based robotic control problems, and enables a real WidowX robotic arm to insert an Ethernet cable into new locations given a sparse task completion signal after only $8$ hours of real world meta-training \footnote{ Results and videos at \url{https://sites.google.com/view/meld-lsm/home}}. To our knowledge, MELD is the first meta-RL algorithm trained in a real-world robotic control setting from images.
\blfootnote{ \hskip -1ex * Equal contribution, \textsuperscript{\textdagger} University of California Berkeley,  \textsuperscript{\textdaggerdbl} Stanford University}
\end{abstract}

\keywords{meta-learning, reinforcement learning, robotic manipulation} 

\section{Introduction}
\label{sec:intro}

General purpose autonomous robots must be able to perform a wide variety of tasks and quickly acquire new skills.
For example, consider a robot tasked with assembling electronics in a data center. 
This robot must be able to insert cables of varying shapes, sizes, colors, and weights into the correct ports with the appropriate amounts of force.
While standard RL algorithms require hundreds or thousands of trials to learn a policy for each new setting, meta-RL methods hold the promise of drastically reducing the number of trials required.
Given a task distribution, such as the variety of ways to insert cables described above, meta-RL algorithms leverage a set of training tasks to meta-learn a mechanism that can quickly learn unseen tasks from the same distribution.
Despite promising results in simulation demonstrating that agents can learn new tasks in a handful of trials~\cite{wang2016learning, duan2016rl,finn2017model,rakelly2019efficient}, these algorithms remain largely unproven on real-world robotic systems.

Applying these algorithms to real-world robotic systems requires handling the raw sensory observations collected by a robot's on-board sensors.
In principle, deep reinforcement learning (RL) algorithms can directly map sensory inputs to actions.
However, this automation comes at a steep cost in sample efficiency since the agent must learn to interpret observations from reward supervision alone.
Fortunately, unsupervised learning of general-purpose latent state (or dynamics) models can serve as an additional training signal to help solve the representation learning problem~\citep{finn2016deep, ghadirzadeh2017deep, lee2019slac, zhang2019solar}.
In this work, we seek to leverage the benefits of latent state models for representation learning to design a meta-RL algorithm that can acquire new skills quickly in the real world.

While meta-learning algorithms are often viewed as algorithms that learn to learn~\cite{schmidhuber1987evolutionary, thrun2005probabilistic, finn2017model}, an alternative viewpoint frames meta-learning as task inference~\cite{tenenbaum1999bayesian, fe2003bayesian, rakelly2019efficient}.
From this perspective, the task is a hidden variable that can be inferred from experience consisting of observations and rewards.
Our key insight is that the same latent dynamics models that greatly improve efficiency in end-to-end single-task RL can \emph{also}, with minimal modification, be used for meta-RL by treating the unknown task information as a hidden variable to be estimated from experience.
We formalize the connection between latent state inference and meta-RL, and leverage this insight in our proposed algorithm \MethodNoSpace, Meta-RL with Latent Dynamics.
To derive MELD, we cast meta-RL and latent state inference into a single  partially observed Markov decision process (POMDP) in which task and state variables are aspects of a more general per-time step hidden variable. 
Concretely, we represent the agent's belief over the hidden variable as the variational posterior in a sequential VAE latent state model that takes observations and rewards as input, and we condition the agent's policy on this belief.
During meta-training, the latent state model and policy are trained across a fixed set of training tasks sampled from the task distribution.
The trained system can then quickly learn a new task from the distribution by inferring the posterior belief over the hidden variable and executing the conditional meta-learned policy.

\begin{figure}[t]
  \centering\vspace{-0.2in}
  \includegraphics[width=0.99\linewidth]{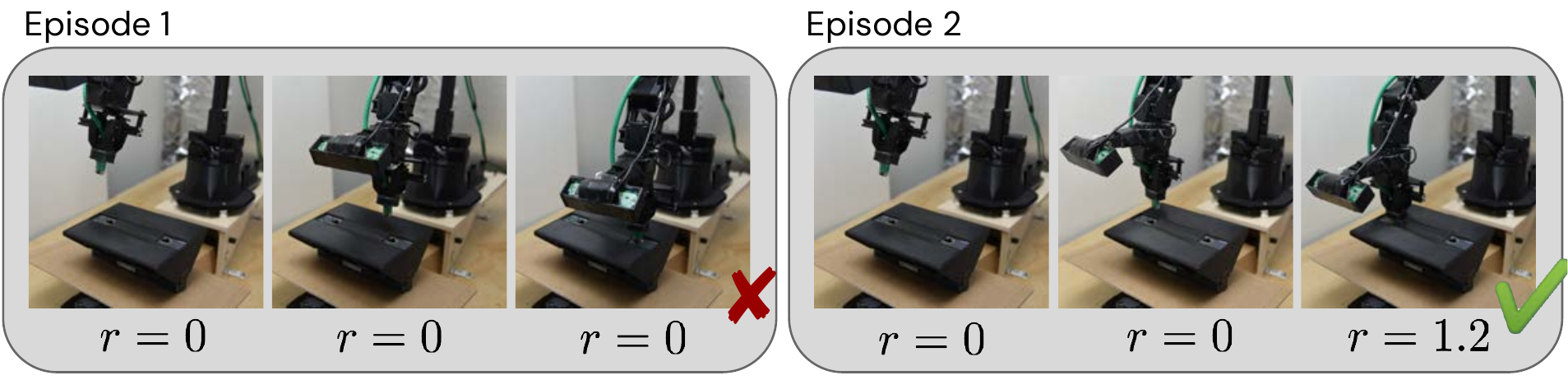}
  \caption{\footnotesize{At test time our algorithm MELD enables a 5-DoF WidowX robot to insert the ethernet cable into a novel insertion location and orientation within two episodes of experience, operating from image observations and a sparse task completion signal when the cable is correctly inserted. 
  MELD achieves this result by meta-training a latent dynamics model to capture task and state information, as well a policy that conditions on this information to explore and identify the correct insertion point.
  }} 
  \label{fig:teaser}
\vspace{-0.2in}
\end{figure}

We find in simulation that \Method substantially outperforms prior work on several challenging locomotion and manipulation problems, such as running at varying velocities, inserting a peg into varying targets, and putting away mugs of unknown weights to varying locations on a shelf. 
We then analyze MELD's capability to meta-learn temporally-extended exploration strategies when only a sparse task completion signal is available.
Finally, using a real WidowX robotic arm, we find that after eight hours of meta-training \Method 
successfully performs Ethernet cable insertion into ports at novel locations and orientations (Figure~\ref{fig:teaser}).
This real-world experiment with the WidowX is to our knowledge the first demonstration of a meta-RL algorithm trained from images on a real robotic platform.
Our open-source implementation of MELD can be found at~\url{https://github.com/tonyzhaozh/meld}.
\vspace{-0.1in}
\section{Related Work}
\label{sec:rw}
Our approach can be viewed methodologically as a bridge between meta-RL methods for fast skill acquisition and latent state models for state estimation.
In this section, we discuss these areas of work, as well as RL on real-world robotic platforms.

\textbf{RL for Robotics.}
While prior work has obtained good results with geometric and force control approaches for a wide range of manipulation tasks~\cite{bicchi2000robotic,pereira2004decentralized,henrich2012robot}, including insertion tasks~\cite{kronander2014task, newman2001interpretation} such as those in our evaluation, such approaches typically require considerable manual design effort for each task.
RL algorithms offer an automated alternative that has been demonstrated on a variety of robotic tasks~\cite{kober2013reinforcement} including insertion~\cite{gullapalli1994acquiring, levine2016end, zeng2018learning, lee2018making, schoettler2019deep}. 
Although these policies learn impressive skills, they typically do not transfer to other tasks and must be re-trained from scratch for each task.

\textbf{Meta-RL.}
\label{sec:rw-meta}
Meta-RL algorithms learn to acquire new skills quickly by using experience from a set of meta-training tasks. 
Current meta-RL methods differ in how this acquisition procedure is represented, ranging from directly representing the learned learning process 
with a recurrent~\cite{wang2016learning, duan2016rl} or recursive~\cite{mishra2017meta-learning} deep network, to learning initial parameters for gradient-based optimization~\cite{finn2017model, rothfuss2018promp, houthooft2018evolved, mendonca2019guided}, and learning tasks via variational inference~\cite{rakelly2019efficient, zintgraf2019varibad, perez2020generalized}.
Some of these works have formalized meta-RL as a special kind of POMDP in which the hidden state is constant throughout a task~\cite{rakelly2019efficient, zintgraf2019varibad, humplik2019meta, perez2020generalized}.
Taking a broader view, we show that meta-RL can be tackled with a general POMDP algorithm that estimates a time-varying hidden state, rendering the same algorithm applicable to problems with both stationary and non-stationary sources of uncertainty.

Within this area, meta-learning approaches that enable few-shot adaptation have been studied with real systems for imitation learning~\cite{finn2017one, yu2018one, james2018task, bonardi2019learning} and goal inference~\cite{xie2018few}, but direct meta-RL in the real world has received comparatively little attention. 
Adapting to different environment parameters has been explored in the sim2real setting for table-top hockey~\cite{arndt2019meta} and legged locomotion~\cite{song2020rapidly}, and in the model-based RL setting for millirobot locomotion~\cite{nagabandi2018learning}.
In Section~\ref{sec:exp-real}, we demonstrate that our algorithm MELD can perform meta-RL trained from images in the real world.

\textbf{Latent State Inference in RL.}
A significant challenge in real-world robotic learning is contending with the complex, high-dimensional, and noisy observations from the robot's sensors.
To handle general partial observability, recurrent policies can persist information over longer time horizons~\cite{yadaiah2006neural, heess2015memory, hausknecht2015deep},  
while explicit state estimation approaches maintain a probabilistic belief over the current state of the agent and update it given experience~\cite{kaelbling1998planning, pineau2003point, ross2011bayesian, deisenroth2012solving, karkus2017qmdp, igl2018deep, gregor2018temporal}.
In our experiments we focus on learning from image observations, which presents a state representation learning challenge that has been studied in detail.
End-to-end deep RL algorithms can learn state representations implicitly, but currently suffer from poor sample efficiency due to the added burden of representation learning~\cite{mnih2013playing, levine2016end, singh34end}.
Pre-trained state estimation systems can predict potentially useful features such as object locations and pose~\cite{tremblay2018deep, kumar2019contextual}; however, these approaches require ground truth supervision.
On the other hand, unsupervised learning techniques can improve sample efficiency without access to additional supervision
~\cite{lange2012autonomous, finn2016deep, schmidt2016self, ghadirzadeh2017deep, florence2019self, yarats2019improving, sax2019mid}. 
Latent dynamics models capture the time-dependence of observations and provide a learned latent space in which RL can be tractably performed~\cite{watter2015embed, karl2016deep, zhang2019solar, hafner2019learning, gelada2019deepmdp, lee2019slac}.
In our work, we generalize the learned latent variable to encode not only 
the state but also the task at hand, enabling efficient meta-RL from images.

\vspace{-.05in}
\section{Preliminaries: Meta-RL and Latent State Models}
\label{sec:method-prob}
In this work, we leverage tools from latent state modeling to design an efficient meta-RL method that can operate in the real world from image observations.
In this section, we review latent state models and meta-RL, developing a formalism that will allow us to derive our algorithm in Section~\ref{sec:method-meld}.

\subsection{POMDPs and Latent State Models}
\label{sec:latent}
We first define the RL problem. A Markov decision process (MDP) consists of a set of states $\mathcal{S}$, a set of actions $\mathcal{A}$, an initial state distribution $p(\tstate_1)$, a state transition distribution $p(\tstate_{t+1}|\tstate_t,\act_t)$, a discount factor $\gamma$, and a reward function $r(\tstate_t, \act_t)$.
We assume the transition and reward functions are unknown, but can be sampled by taking actions in the environment.
The goal of RL is to learn a policy $\pi(\act_t|\tstate_t)$ that selects actions that maximize the sum of discounted rewards. However, robots operating in the real world do not have access to the underlying state $\tstate_t$, and must instead select actions using high-dimensional and often incomplete observations from cameras and other sensors. Such a system can be described as a partially observed Markov decision process (POMDP), where observations $\obs_t$ are a noisy or incomplete function of the unknown underlying state $\tstate_t$, and the policy is conditioned on a history of observations as $\pi(\act_t | \obs_{1:t})$.

While end-to-end RL methods can acquire representations of observations from reward supervision alone~\cite{levine2016end, zeng2018learning}, the added burden of end-to-end representation learning limits sample efficiency and can make optimization more difficult. Methods that explicitly address this representation learning problem are more efficient and scale more efficiently to harder domains~\cite{igl2018deep, gelada2019deepmdp, lee2019slac}.
These approaches train latent state models to learn meaningful representations of the incoming observations by explicitly representing the unknown Markovian state as a hidden variable $\state_t$ in a graphical model, as shown in Figure~\ref{fig:pgm} (a).
The parameters of these graphical models can be trained by approximately maximizing the log-likelihood of the observations: $\log p(\obs_{1:T} | \act_{1:T-1}) = \log \int p(\obs_T | \state_T) p(\state_T | \state_{T-1}, \act_{T-1}) ... p(\state_1)dz$.
Given a history of observations and actions seen so far, the posterior distribution over the hidden variable captures the agent's belief over the current underlying state, and can be written as $\be_t = p(\state_t | \obs_{1:t}, \act_{1:t-1})$.
Then, rather than conditioning the policy on raw observations, these methods learn a policy $\pi(\act_t|\be_t)$ as a function of this belief state.

\subsection{Meta-Reinforcement Learning}
\label{sec:meta}
In this work, we would like a robot to learn to acquire new skills quickly. 
We formalize this problem statement as meta-RL, where each task $\task$ from a distribution of tasks $p(\task)$ is a POMDP as described above, with initial state distribution $p_\task(\tstate_1)$, dynamics function $p_\task(\tstate_{t+1}|\tstate_t,\act_t)$, observation function $p_\task(\obs_t | \tstate_t)$, and reward function $r_\task(r_t|\tstate_t,\act_t)$, as shown in Figure~\ref{fig:pgm} (b).
For example, a task distribution that varies both dynamics and rewards across tasks may consist of placing mugs of varying weights (dynamics) in different locations (rewards) on a kitchen shelf.
The meta-training process leverages a set of training tasks sampled from $p(\task)$ to learn an adaptation procedure that can adapt to a new task from $p(\task)$ using a small amount of experience. 
Similar to the framework in probabilistic inference-based and recurrence-based meta-RL approaches~\cite{zintgraf2019varibad, rakelly2019efficient, duan2016rl, wang2016learning}, we formalize the adaptation procedure $f_\phi$ as a function of experience $( \obs_{1:t}, r_{1:t}, \act_{1:t-1} )$ that summarizes task-relevant information into the variable $\mathbf{c}_t$.
The policy is conditioned on this updated variable as $\pi_\theta(\act_t|\obs_t, \mathbf{c}_t)$ to adapt to the task. 
By training the adaptation mechanism $f_\phi$ and the policy $\pi_\theta$ end-to-end to maximize returns of the adapted policy, meta-RL algorithms can learn policies that effectively modulate and adapt their behavior with small amounts of experience in new tasks. 
We formalize this meta-RL objective as:
\begin{align}\vspace{-0.2in}
    \max_{\theta,\phi} \hspace{4pt} \mathop{\mathbb{E}}_{\task \sim p(\task)} \hspace{4pt}
    \mathop{\mathbb{E}}_{\substack{
    \obs_{t} \sim p_\task(\cdot|\tstate_t) 
    \\ \act_t \sim \pi_\theta(\cdot|\obs_t,\mathbf{c}_t) 
    \\ \tstate_{t+1} \sim p_\task(\cdot|\tstate_t,\act_t) 
    \\ r_t \sim r_\task(\cdot | \tstate_t, \act_t)
    }} \hspace{4pt}
    \left[ \sum_{t=1}^T  \gamma^t r_t \right]
    \hspace{10pt} \text{where} \hspace{10pt} \mathbf{c}_t = f_{\phi}(\obs_{1:t}, r_{1:t}, \act_{1:t-1}).
    \label{eqn:meta-general}
    \vspace{-0.3in}
\end{align}
Meta-RL methods may differ in how the adaptation procedure $f_\phi$ is represented (e.g., as probabilistic inference~\cite{rakelly2019efficient, zintgraf2019varibad}, as a recurrent update~\cite{duan2016rl, wang2016learning}, as a gradient step~\cite{finn2017model}), how often the adaptation procedure occurs (e.g., at every timestep~\cite{duan2016rl, zintgraf2019varibad} or once per episode~\cite{rakelly2019efficient, humplik2019meta}), and also in how the optimization is performed (e.g., on-policy~\cite{duan2016rl}, off-policy~\cite{rakelly2019efficient}). 
Differences aside, these methods all typically optimize this objective end-to-end, creating a representation learning bottleneck when learning from image inputs that are ubiquitous in real-world robotics. 
In the following section, we show how the latent state models discussed in Section~\ref{sec:latent} can be re-purposed for joint representation and task learning, and how this insight leads to a practical algorithm for image-based meta-RL.

\section{\Method}
\label{sec:method-meld}
In this section, we present \MethodNoSpace: an efficient algorithm for meta-RL from images. We first develop the algorithm in Section~\ref{sec:alg} and then describe its implementation in Section~\ref{sec:imp}.

\subsection{Meta-RL with Latent Dynamics Models} 
\label{sec:alg}

\begin{figure}
  \centering
    \includegraphics[width=\textwidth]{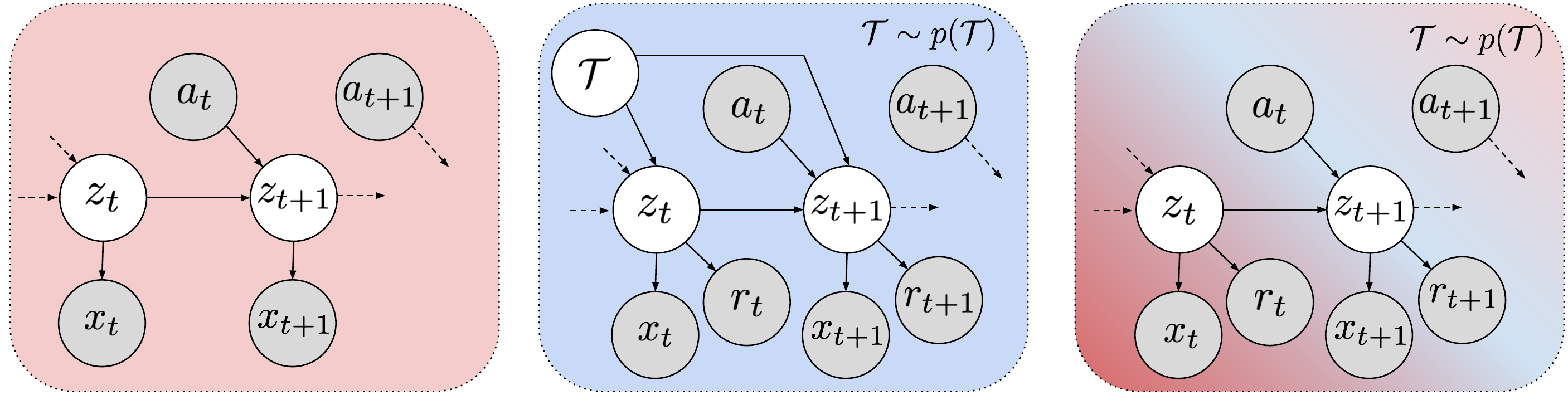}
      \caption {\footnotesize{\textbf{(a)}: When only partial observations of the underlying state are available, latent dynamics models can glean state information $\z_t$ from a history of observations.
      \textbf{(b)}: Meta-RL considers a task distribution where the current task $\mathcal{T}$ is an unobserved variable that controls dynamics and rewards. \textbf{(c)}: We interpret $\mathcal{T}$ as part of $\z_t$, allowing us to leverage latent dynamics models for efficient image-based meta-RL.}
  \label{fig:pgm}
  }\vspace{-0.1in}
\end{figure}

To see how task inference in meta-RL can be cast as latent state inference, consider the graphical models depicted in Figure~\ref{fig:pgm}. Panel (a)
illustrates a standard POMDP with underlying latent state $\z_t$ and observations $\obs_t$, and panel (b) depicts standard meta-RL, where the hidden task variable $\task$ is assumed constant throughout the episode.
In the meta-RL setting, the policy must then be conditioned on both the observation and the task variable in order to adapt to a new task (see Equation~\ref{eqn:meta-general}).
Casting this task variable as part of the latent state, panel (c) illustrates our graphical model, where the states $\state_t$ now contain both state and task information. In effect, we cast the task distribution over POMDPs as a POMDP itself, where the state variables now additionally capture task information. This meld allows us to draw on the rich literature of latent state models discussed in Section~\ref{sec:latent}, and use them here to tackle the problem of meta-RL from sensory observations. 
Note that the task is not explicitly handled since it is simply another hidden state variable, providing a seamless integration of meta-RL with learning from sensory observations.
\definecolor{meldcolor}{RGB}{194, 64, 64}
\renewcommand\algorithmiccomment[1]{\hfill $\triangleright$ \textcolor{meldcolor}{#1}}

\newcommand\mycommfont[1]{\footnotesize\ttfamily\textcolor{blue}{#1}}
\SetCommentSty{mycommfont}
\begin{wrapfigure}{r}{0.49\textwidth} \vspace{-3.0in}
\begin{minipage}{0.5\textwidth}

\begin{algorithm}[H]
\caption{{\bf MELD Meta-training}}
\begin{algorithmic}[1]
\REQUIRE Training tasks $\{\task_i\}_{i=1 \ldots J}$ from $p(\task)$ , learning rates $\eta_1, \eta_2, \eta_3$ 
\STATE Init. model $p_{\phi}$, $q_{\phi}$, actor $\pi_{\theta}$, critic $Q_{\zeta}$
\STATE Init. replay buffers $\mathcal{B}_i$ for each train task
\WHILE{not done}
\FOR[collect data]{each task $\task_i$}
\STATE Infer $\be_1 = q_\phi(\state_1|\obs_1, r_1)$
\STATE Step with $\act_1 \sim \pi_{\theta}(\act | \be_1)$, get $\obs_{2}$, $r_{2}$
\FOR{$t = 2, \ldots T-1$}
\STATE Infer $\be_t = q_{\phi}(\z_t | \obs_t, r_t, \z_{t-1}, \act_{t-1})$ 
\STATE Step $\act_t \sim \pi_{\theta}(\act | \be_t)$, get $\obs_{t+1}$, $r_{t+1}$
\ENDFOR
\STATE Add data $\{ \obs_{1:T}, r_{1:T}, \act_{1:T-1} \}$ to $\mathcal{B}^{i}$
\ENDFOR
\FOR{step in train steps}
\FOR{each task $\task_i$}
\STATE Sample $\{ \obs_{1:T}, r_{1:T}, \act_{1:T-1} \} \sim \mathcal{B}_i$
\STATE Infer beliefs $\be_{1:T} = \{q_{\phi} (\state_t | \dots)\}_{1:T}$ 
\STATE Predict reconstructions $\{\hat{\obs}_t, \hat{r}_t\}_{1:T}$
\STATE  $\mathcal{L}^i_{m} = \mathcal{L}_{model}(\{\obs_t, \hat{\obs}_t, r_t, \hat{r}_t, \act_t\}_{1:T})$
\ENDFOR
\STATE  $\phi \gets \phi - \eta_1 \nabla_\phi \sum_i \mathcal{L}^i_{m}$ \hfill $\triangleright$ \textcolor{meldcolor}{train model}
\STATE Update $\theta, \zeta$ with SAC($\eta_2, \eta_3$) \hfill $\triangleright$ \textcolor{meldcolor}{train AC}
\ENDFOR
\ENDWHILE
\end{algorithmic}
\label{alg:meta-train}
\end{algorithm}
\vspace{-0.25in}


\begin{algorithm}[H]
\caption{{\bf MELD Meta-testing}}
\begin{algorithmic}[1]
\REQUIRE Test task $\task \sim p(\task)$
\STATE Infer $\be_1 = q_{\phi}(\z_1|\obs_1, r_1)$
\STATE Step with $\act_1 \sim \pi_{\theta}(\act | \be_1)$, get $\obs_{2}$, $r_{2}$
\FOR{$t=2, \ldots, T-1$}
\STATE Infer $\be_t = q_{\phi}(\z_t | \obs_t, r_t, \z_{t-1}, \act_{t-1})$
\STATE Step with $\act_t \sim \pi_{\theta}(\act | \be_t)$, get $\obs_{t+1}$, $r_{t+1}$
\ENDFOR
\end{algorithmic}
\label{alg:meta-test}
\end{algorithm}
\end{minipage}\vspace{-3in}

\end{wrapfigure}


\begin{minipage}[h]{0.49\textwidth}
Concretely, we learn a latent state model over hidden 
variables by optimizing 
the log-likelihood of the evidence (observations and rewards) in the graphical model in Figure~\ref{fig:pgm}c: 
\begin{align}
    \max_{\phi}  \hspace{-4pt} \mathop{\mathbb{E}}_{\task \sim p(\task)} 
    \hspace{-6pt}
    \mathop{\mathbb{E}}_{\substack{\obs_t \sim p_\task(\cdot | \tstate_t)
    \\ \act_t \sim \pi_\theta(\cdot|\be_t)
    \\ \tstate_{t+1} \sim p_{\task}(\cdot|\tstate_t,\act_t) 
    \\ r_t \sim r_{\task}(\cdot | \tstate_t, \act_t) 
    }} \hspace{-12pt}
    \left[ \log p_{\phi}(\obs_{1:T}, r_{1:T} | \act_{1:T-1}) \right].
    \label{eqn:meta-model} \vspace{-0.4in}
\end{align}
Note that the only change from the latent state model from Section~\ref{sec:latent} is the inclusion of rewards as part of the observed evidence.
While this change appears simple, it enables meta-learning by allowing the hidden state to capture task information.
Posterior inference in this model then gives the agent's belief $\be_t = p(\state_t | \obs_{1:t}, r_{1:t}, \act_{1:t-1})$ over latent state and task variables $\state_t$.
Conditioned on this belief, the policy $\pi_\theta(\act_t|\be_t)$ can learn to adapt its behavior to the task. 
Prescribing the adaptation procedure $f_\phi$ from Equation~\ref{eqn:meta-general} to be posterior inference in our latent state model,
the meta-training objective in \Method is: \\
\vspace{-0.2in}
\begin{align}
    \max_{\theta} \hspace{4pt} \mathop{\mathbb{E}}_{\substack{\task \sim p(\task)}} \hspace{4pt}
    \mathop{\mathbb{E}}_{\substack{\obs_t \sim p_\task(\cdot | \tstate_t)
    \\ \act_t \sim \pi_\theta(\cdot|\be_t) 
    \\  \tstate_{t+1} \sim p_{\task}(\cdot|\tstate_t,\act_t) 
    \\  r_t \sim r_{\task}(\cdot | \tstate_t, \act_t)
    }} \hspace{4pt}
    \left[ \sum_{t=1}^T  \gamma^t r_t \right]
    \\
    \hspace{-30pt} \text{where} \hspace{5pt} \be_t = p(\state_t|\obs_{1:t},r_{1:t},\act_{1:t-1}).
    \label{eqn:meld-rl}
    \vspace{-0.5in}
\end{align}
\hfill \\
By melding state and task inference, \Method inherits the same representation learning mechanism as latent state models discussed in Section~\ref{sec:latent} to enable efficient meta-RL with images.
\end{minipage}

\vspace{-2pt}
\subsection{Implementing \Method} 
\label{sec:imp}
\vspace{-6pt}
\begin{minipage}[h]{1.0\textwidth}
Exactly computing the posterior distribution over the latent state variable is intractable, so we take a variational inference approach to maximize a lower bound on the log-likelihood objective~\cite{wainwright2008graphical} in Equation~\ref{eqn:meta-model}. 
We factorize the variational posterior as $q(\state_{1:T} | \obs_{1:T}, r_{1:T}, \act_{1:T-1}) = q(\state_T | \obs_T, r_T, \state_{T-1}, \act_{T-1}) \dots q(\state_2 | \obs_2, r_2, \state_1, \act_1) q(\state_1 | \obs_1, r_1)$. 
With this factorization, we implement each component as a deep neural network and optimize the evidence lower bound of the joint objective, where $\mathop{\E}_{\state_{1:t} \sim q_{\phi}} \left[ \log p(\obs_{1:T}, r_{1:T} | \act_{1:T-1}) \right] \geq \mathcal{L}_{model}$, with $\mathcal{L}_{model}$
defined as:
\vspace{-10pt}
\begin{multline}
    \mathcal{L}_{model}(\obs_{1:T}, \textcolor{blue}{r_{1:T}}, \act_{1:T-1}) =  \mathop{\E}_{\state_{1:T} \sim q_\phi} \sum_{t=1}^{T} \log p_\phi(\obs_{t} | \state_{t}) + \log p_\phi(\textcolor{blue}{r_{t}} | \state_{t}) \\ -  D_\text{KL}(q_\phi(\state_1 | \obs_1, \textcolor{blue}{r_1}) \| p(\state_1))  - \sum_{t=2}^{T} D_\text{KL}(q_\phi(\state_{t} | \obs_{t}, \textcolor{blue}{r_{t}}, \state_{t-1}, \act_{t-1}) \| p_\phi(\state_{t} | \state_{t-1}, \act_{t-1})).
\label{eq:elbo}\vspace{-0.3in}
\end{multline}
The first two terms encourage a rich latent representation $\state_t$ by requiring reconstruction, while the last term keeps the inference network consistent with latent dynamics. 
The first timestep posterior $q_\phi(\state_1 | \obs_1, r_1)$ is modeled separately from the remaining steps, and $p(\state_1)$ is chosen to be a fixed unit Gaussian $\mathcal{N}(0,I)$. 
The learned inference networks $q_\phi(\state_1|\obs_1,r_1)$ and $q_\phi(\state_t|\obs_{t}, r_{t}, \state_{t-1}, \act_{t-1})$, decoder networks $p_\phi(\obs_t|\state_t)$ and $p_\phi(r_t|\state_t)$, and dynamics $p_\phi(\z_t | \z_{t-1}, \act_{t-1})$ are all fully connected networks that output parameters of Gaussian distributions.
We follow the architecture of the latent variable model from SLAC~\cite{lee2019slac} and provide the remaining implementation details in Appendix~\ref{sec:app-impl}.
\end{minipage}

\begin{wrapfigure}{r}{0.59\textwidth}
\begin{minipage}[h]{0.59\textwidth}
    \centering
    \vspace{-30mm}
    \includegraphics[width=0.99\linewidth]{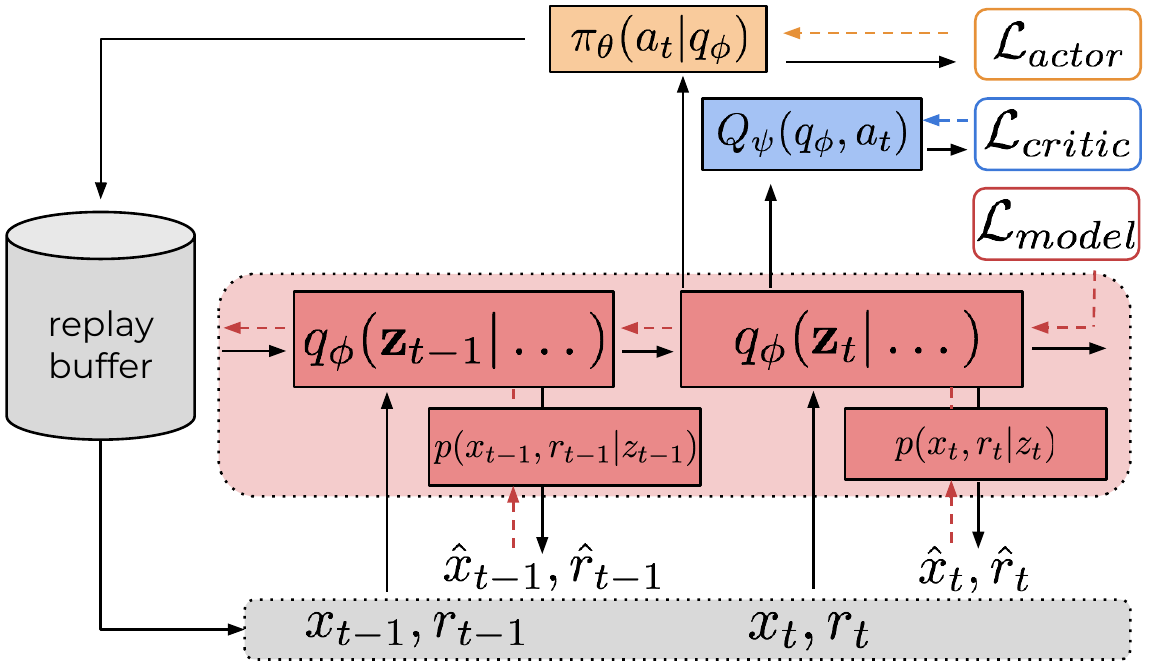}
    \captionof{figure}{\footnotesize{MELD meta-training alternates between collecting data with $\pi_{\theta}$ and training the latent state model $p_{\phi}$, inference networks $q_{\phi}$, actor $\pi_{\theta}$, and critic $\mathcal{Q}_{\zeta}$.}
    }
    \label{fig:training}\vspace{-0.2in}
\end{minipage}
\end{wrapfigure}

\begin{minipage}[h]{0.39\textwidth}
We use the soft actor-critic (SAC)~\cite{haarnoja2018soft} RL algorithm in this work, due to its high sample efficiency and performance.
The actor $\pi_\theta(\act_t|\be_t)$ and the critic $Q_\psi(\be_t, \act_t)$ are conditioned on the posterior belief $\be_t$, modeled as fully connected neural networks, and trained as prescribed by the SAC algorithm.
During meta-training, \Method alternates between collecting data with the current policy, training the model by optimizing $\mathcal{L}_{model}$, and training the policy with the current model. 
Meta-training and meta-testing are described in Algorithm~\ref{alg:meta-train} and~\ref{alg:meta-test} respectively. 
\end{minipage}

\vspace{-0.1in}
\section{Experiments}\label{sec:experiments}

\begin{minipage}[h]{1.0\textwidth}
\noindent In our experiments, we aim to answer the following: 
\textbf{(1)} How does \Method compare to prior meta-RL methods in enabling fast acquisition of new skills at test time in challenging simulated control problems?
\textbf{(2)} Can \Method meta-learn effective exploration when only sparse task completion rewards are available at meta-test time?
\textbf{(3)} Can \Method enable real robots to quickly acquire skills via meta-RL from images?

\vspace{0.1in}
\subsection{MELD in Simulated Environments}
\label{sec:exp-sim}
\end{minipage}

\begin{wrapfigure}{R}{0.4\textwidth}
\begin{minipage}{0.4\textwidth}
  \centering\vspace{-0.15in}
  \includegraphics[height=0.12\textheight]{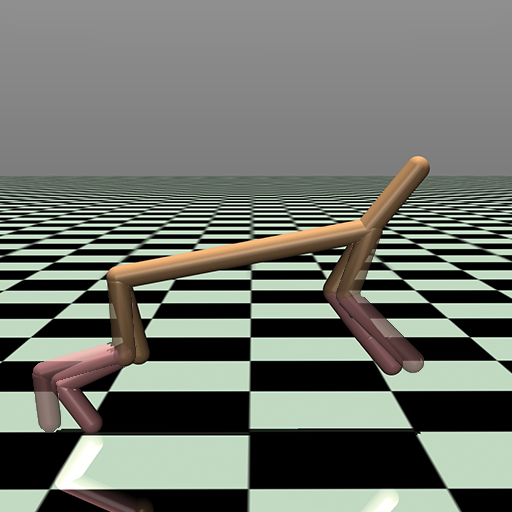}
    \includegraphics[height=0.12\textheight]{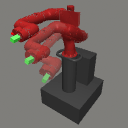}
    \includegraphics[height=0.12\textheight]{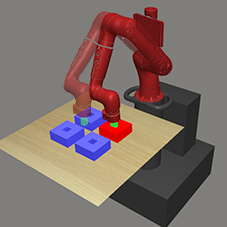} 
    \includegraphics[height=0.12\textheight]{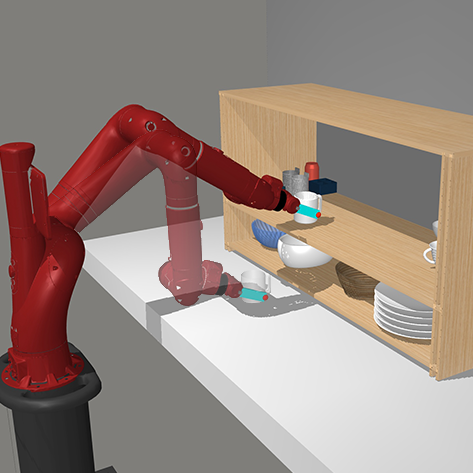}
  \captionof{figure}{\footnotesize{Simulated locomotion and manipulation meta-RL environments in the MuJoCo simulator~\citep{todorov2012mujoco}; goals illustrated for visualization purposes.} 
  }
  \vspace{-10pt}
  \label{fig:sim-tasks}\vspace{4pt}
\end{minipage}
\end{wrapfigure}

In this section, we evaluate \Method on the four simulated image-based continuous control problems in Figure~\ref{fig:sim-tasks}. 
In (a) Cheetah-vel, each task is a different target running velocity for the 6-DoF legged robot. 
Reward is the difference in robot velocity from the target.
The remainder of the problems use a 7-DOF Sawyer robotic arm. 
In (b) Reacher, each task is a different goal position for the end-effector.
In (c) Peg-insertion, the robot must insert the peg into the correct box, where each task varies the goal box as well as locations of all four boxes. 
In (d) Shelf-placing, each task varies the weight (dynamics change) and target location (reward change) of a mug that the robot must move to the shelf.
For the Sawyer environments, the reward function is the negative distance between the robot end-effector and the desired end location.
For all environments, we train with $30$ meta-training tasks and evaluate on $10$ meta-test tasks from the same distribution that are not seen during training.

We compare MELD to two representative state-of-the-art meta-RL algorithms, PEARL~\cite{rakelly2019efficient} and RL$^2$~\cite{duan2016rl}. 
PEARL models a belief over a probabilistic latent task variable as a function of un-ordered batches of transitions, and conditions the policy on both the current observation and this inferred task belief.
Unlike MELD, this algorithm assumes an exploration phase of several trajectories in the new task to gather information before adapting, so to get its best performance, we evaluate only \emph{after} this exploration phase.
RL$^2$ models the policy as a recurrent network that directly maps observations, actions, and rewards to actions. 
To apply PEARL and RL$^2$ with image observations, we augment them with the same convolutional encoder architecture used by \MethodNoSpace. 
Finally, to verify the need for task inference to solve new tasks, we compare to SLAC~\cite{lee2019slac}, which infers state information from a sequence of observations but does \emph{not} perform meta-learning.

\begin{figure}
  \centering\vspace{-.4in} 
    \includegraphics[height=0.135\textheight]{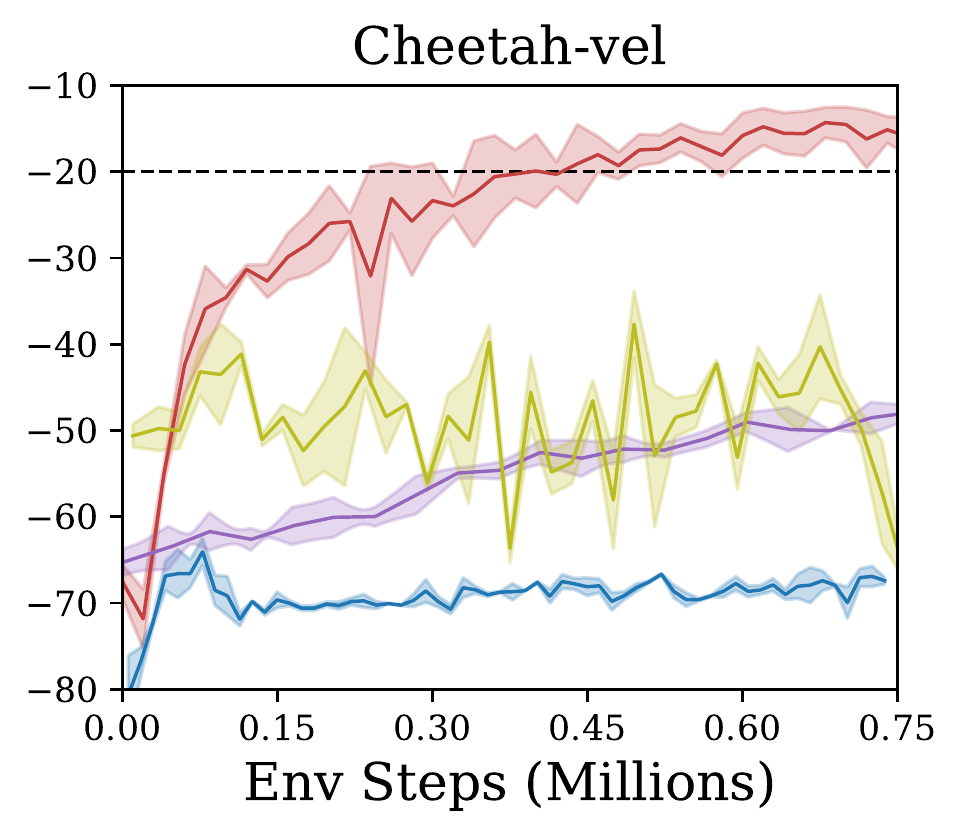} \hskip -2ex
    \includegraphics[height=0.135\textheight]{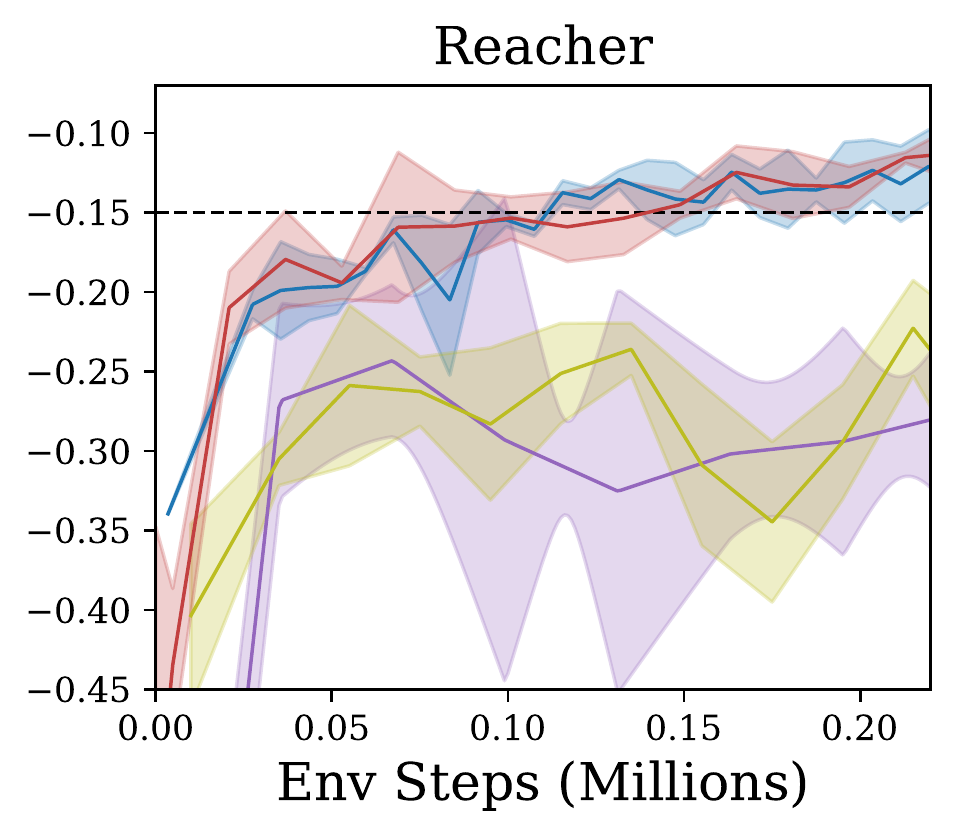} \hskip -1ex
    \includegraphics[height=0.135\textheight]{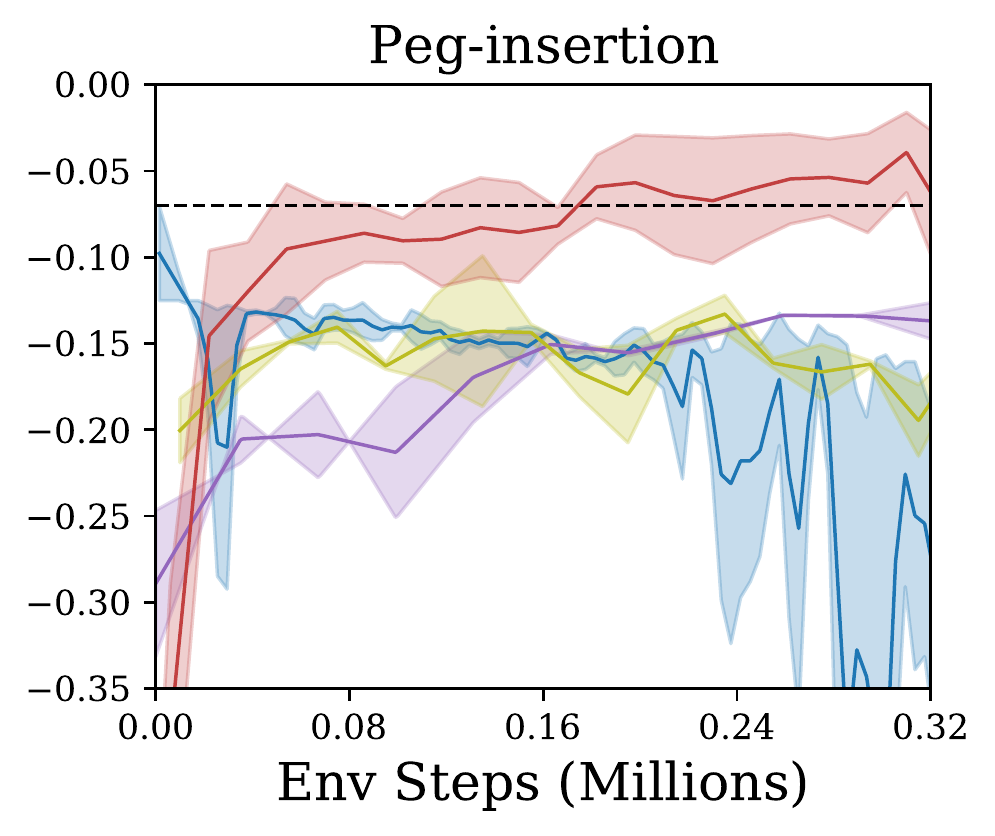} \hskip -2ex
    \includegraphics[height=0.135\textheight]{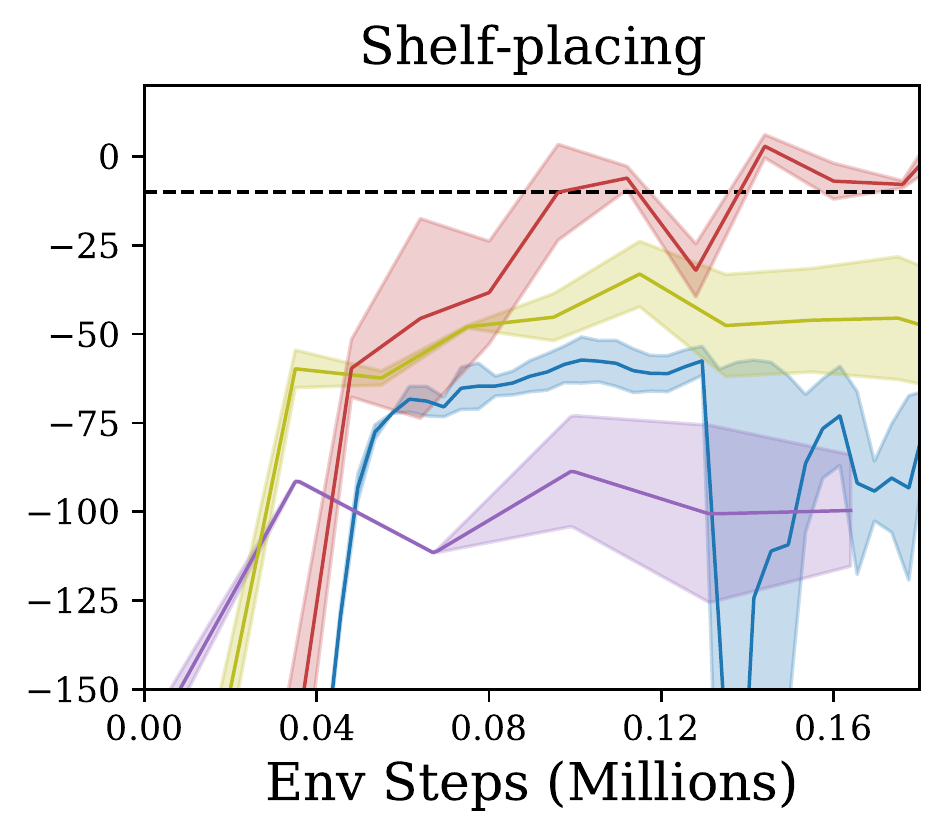}
    \includegraphics[width=0.8\textwidth]{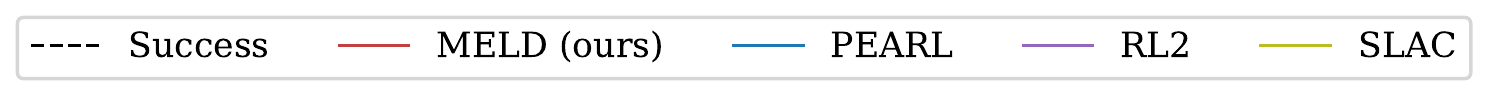}\vspace{-0.1in}
  \caption{\footnotesize{Rewards on test tasks versus meta-training environment steps, comparing MELD to prior methods. Black dashed line indicates consistent task success (see Appendix~\ref{sec:app-sim} for definition).}
  }
  \label{fig:exp-comparisons}
  \vspace{-0.15in}
\end{figure}

In Figure~\ref{fig:exp-comparisons} we plot average performance on meta-test tasks over the course of meta-training, across 3 random seeds.
See Appendix~\ref{sec:app-sim} for the the definitions of metrics used for each task.
\Method achieves the highest performance in each environment and is the only method to fully solve Cheetah-vel, Peg-insertion, and Shelf-placing.
The SLAC baseline fails in this meta-RL setting, as expected, with qualitative behavior of always executing a single ``average'' motion, such as reaching toward a mean goal location and running at a medium speed.
PEARL aggregates task information over time in its latent task variable, but relies on the current observation alone for state information.
Its poor performance on Cheetah, Reacher, and Shelf reflect the need for state estimation from a sequence of observations to perform control in these environments.
While RL$^2$ is capable of propagating both state and task information over time, we observe that it overfits heavily to training tasks and struggles on evaluation tasks.

\subsection{Temporally-Extended Exploration}
\label{sec:exp-button}

\begin{wrapfigure}{r}{0.55\textwidth} 
\centering \vspace{-0.10in}
\includegraphics[height=0.20\textheight]{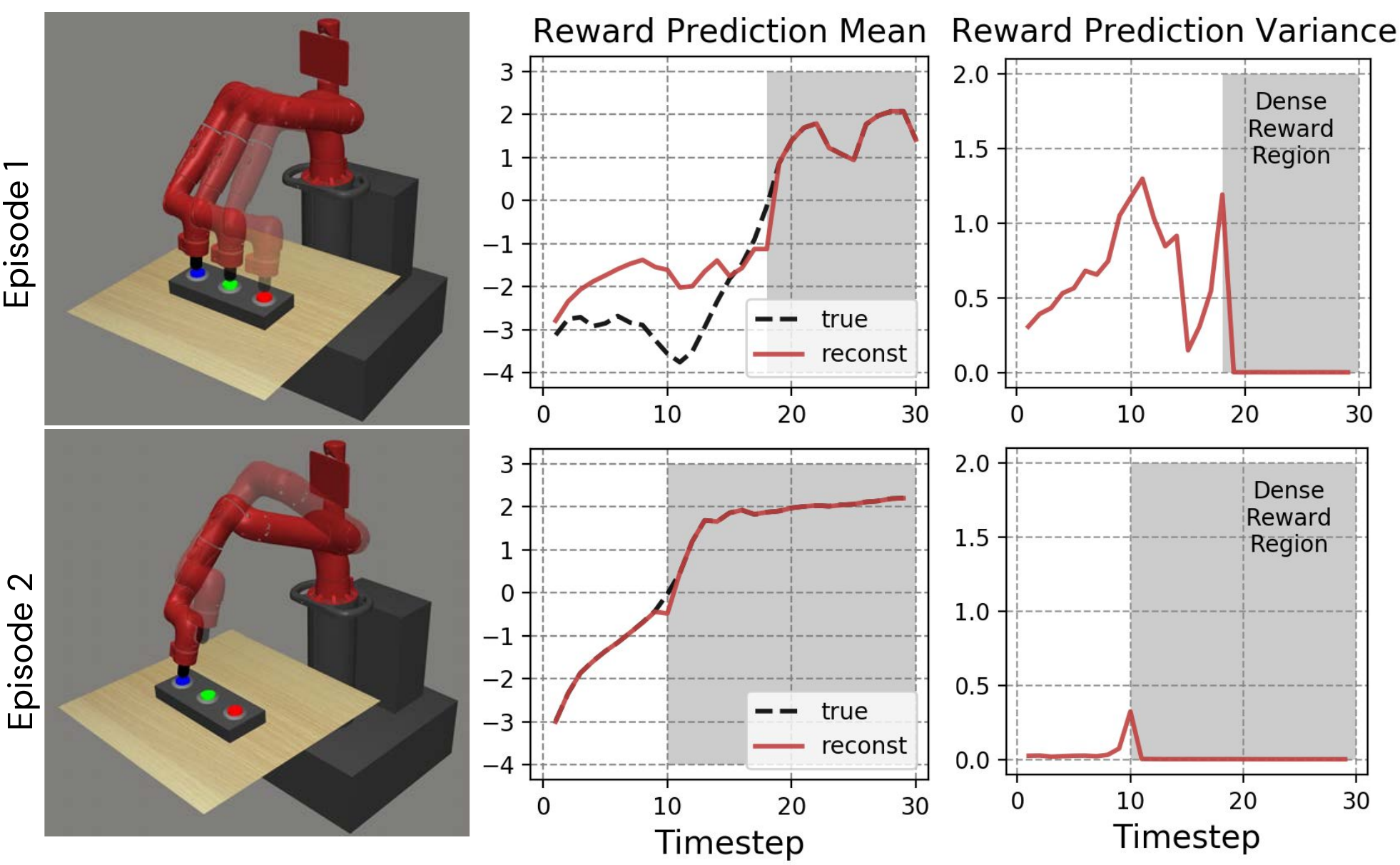}
  \captionof{figure}{
  \footnotesize{Button press with reward given only upon pressing correct button. The robot explores each button until it finds the correct one (top) and returns to that button immediately in the next episode (bottom). See text for discussion.}
  }
  \vspace{-0.05in}
  \label{fig:exp-button} 
\end{wrapfigure}

The previous section assumed a shaped reward function that is the negative distance between the current robot position and the desired one at every timestep.
In the real world, this type of reward function is typically not available to the agent, since it requires information that may be difficult or impossible to obtain.
For example, in the Ethernet cable insertion problem, the location of the insertion point is unknown, but the agent might receive a sparse task completion reward upon making the correct electrical connection.
To quickly succeed at a new task given only this sparse signal at meta-test time, it is critical for MELD to reason over multiple episodes and acquire temporally-extended strategies during meta-training.
Because RL with sparse rewards is very inefficient, we follow prior work~\cite{gupta2018meta, rakelly2019efficient} and assume access to a shaped reward function during meta-training to help learn these strategies. 
We detail our particular approach to making use of shaped rewards during meta-training in Appendix~\ref{sec:app-sparse}, and at meta-test time assume access to only the sparse reward signal.

\begin{wrapfigure}{r}{0.25\textwidth}
\centering
\vspace{-0.14in} 
\includegraphics[trim={0.5cm 0.1cm 0 0}, width=0.26\textwidth]{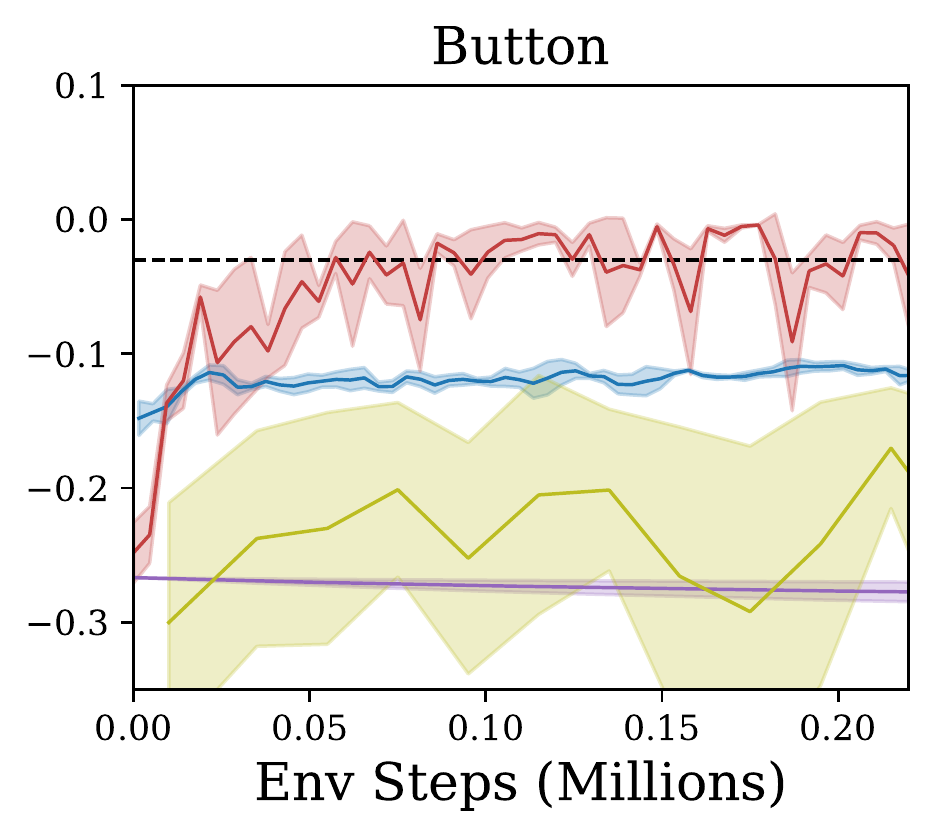} 
\vspace{-0.22in}
  \captionof{figure}{
  \footnotesize{Comparison to approaches from Figure~\ref{fig:exp-comparisons} on the button-pressing environment, showing test task performance versus meta-training environment steps.}
  }
  \vspace{-0.1in}
  \label{fig:exp-button-score}
\end{wrapfigure}

To evaluate MELD in this setting, we design a simulated button-pressing environment with the Sawyer robot, where the button to push and the location of the panel changes with each task. 
Sparse reward is given only when the correct button is pushed, while shaped reward (used only in meta-training) is the negative distance from the robot's end-effector to the insertion point.
In Figure~\ref{fig:exp-button}, we analyze the qualitative behavior of MELD when learning a new task at test time.
Though the shaped reward is not used at test time, we plot the shaped reward reconstruction mean and variance to gain an insight into the contents of the learned latent state.
In the first episode, predicted reward error and variance are high until the robot presses the correct button.
MELD's latent state model persists the task information to the second episode (predicted reward error and variance are very low) and the robot navigates immediately to the correct button.
In Figure~\ref{fig:exp-button-score}, we compare \Method to the same baselines introduced in the previous section and find that it is the only method able to press the correct button within two trajectories of experience.
In the next section, we test MELD's capability to perform such exploration and exploitation in the real world.

\subsection{\Method in the Real World}
\label{sec:exp-real}
\begin{wrapfigure}{r}{0.45\textwidth}
\begin{minipage}[h]{.45\textwidth}
  \centering 
    \includegraphics[height=0.18\textheight]{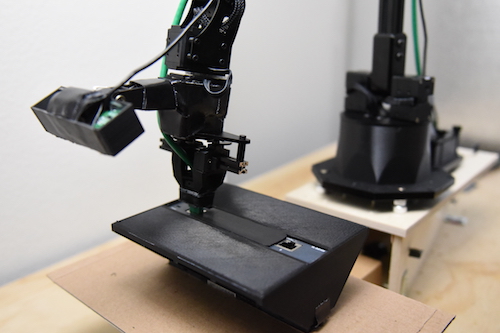}
   \captionof{figure}{\footnotesize{Ethernet cable insertion problem. The automatic reset mechanism changes the position and orientation of the router across tasks.}}
  \label{fig:hardware-task}
  \centering
  \vspace{0.15in}
    \includegraphics[height=0.13\textheight]{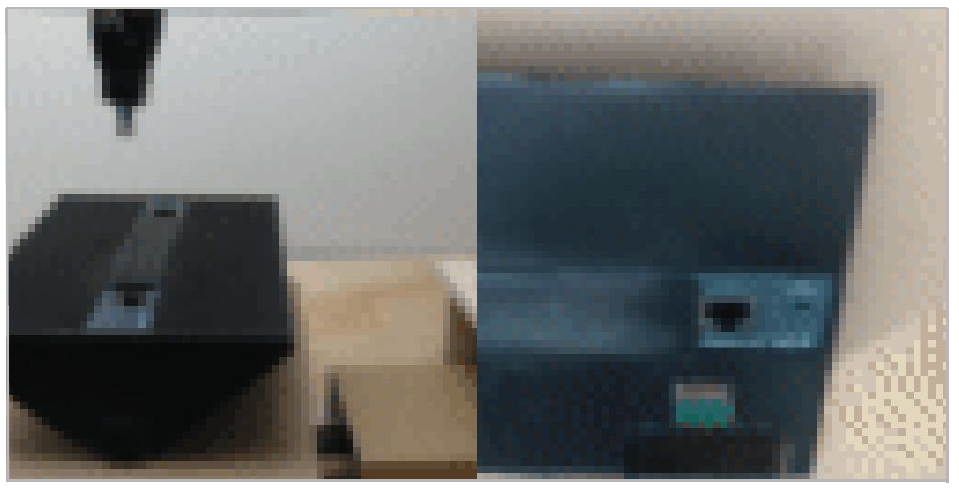}
  \captionof{figure}{\footnotesize{$64$x$128$ image observations seen by WidowX.}}
  \label{fig:hardware-obs}
  \vspace{-0.15in}
 \end{minipage}
\end{wrapfigure}

We now evaluate \Method on a real-world $5$-DoF WidowX arm performing Ethernet cable insertion. 
The task distribution consists of different ports in a router that also varies in location and orientation (see Figure~\ref{fig:hardware-task}).
To instrument these tasks in the real world, we build an automatic reset mechanism that moves and rotates the router, as detailed in Appendix~\ref{sec:app-widowx}.
At meta-test time the reward is a sparse signal given when the robot inserts the cable in the correct port.
As in the previous section, during meta-training we make use of a shaped reward function that is the sum of the L$2$-norms of translational and rotational distances between the pose of the object in the end-effector and a goal pose.
The agent's observations are concatenated images from two webcams (Figure~\ref{fig:hardware-obs}): one fixed view and one first-person view from a wrist-mounted camera. 
The policy sends joint velocity controls over a ROS interface to a low-level PID controller to move the joints of each robot. 

\begin{wrapfigure}{r}{0.6\textwidth}
\vspace{.1in}
  \centering \vspace{-0.3in}
  \hskip -1ex
    \includegraphics[height=0.14\textheight]{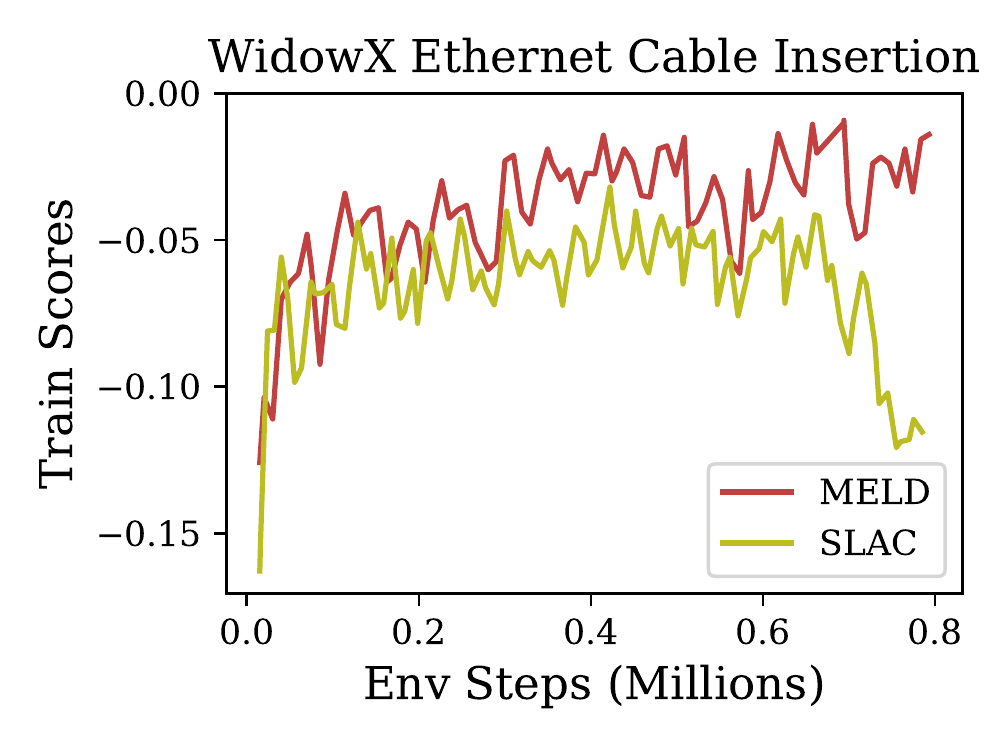} \hskip -1ex
    \includegraphics[height=0.14\textheight]{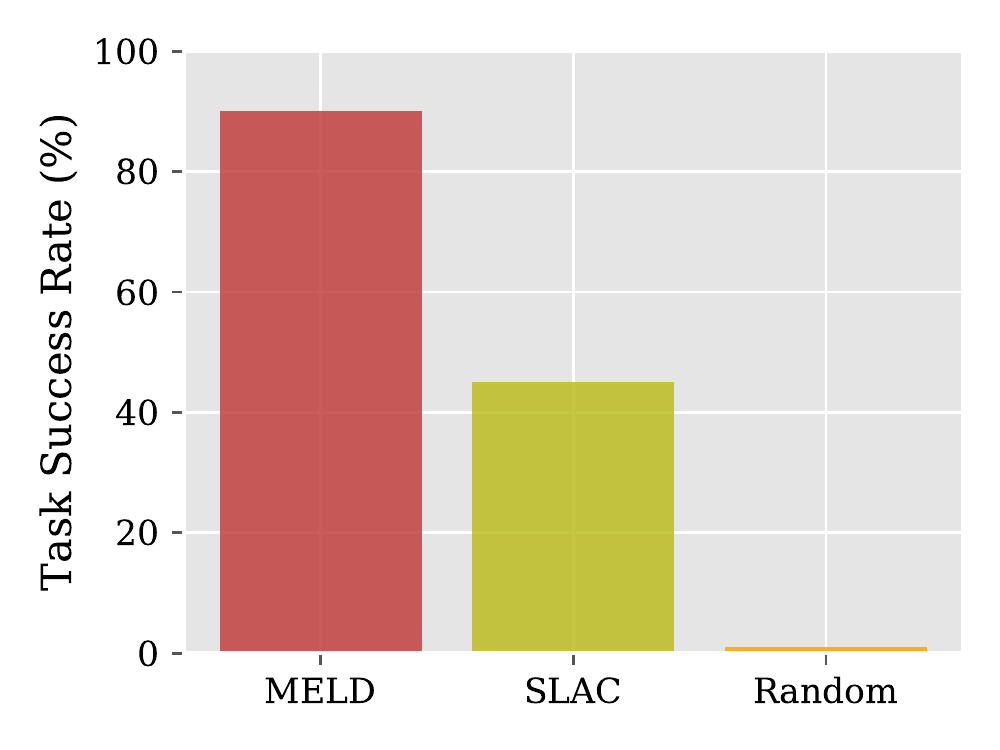} \hskip -1ex
  \caption{\textbf{(left)} Training task rewards versus meta-training environment steps. \textbf{(right)} Task success (defined as full insertion into correct hole) on held-out tasks after meta-training.
  }
  \label{fig:hardware-plot}
  \vspace{-0.15in}

\end{wrapfigure}

We compare \Method to SLAC as described in Section~\ref{sec:exp-sim} as well as a random policy, and plot the results in Figure~\ref{fig:hardware-plot}. 
After training across $20$ meta-training tasks using a total of $8$ hours ($80,000$ samples at $3.3$Hz) worth of data, \Method achieves a success rate of $90\%$ over three rounds of evaluation in each of the $10$ randomly sampled evaluation tasks that were not seen during training. 
To our knowledge, this experiment is the first demonstration of meta-RL trained entirely in the real world from image observations.
We also conducted experiments with the Sawyer robot, finding that MELD enables the Sawyer to insert a peg into the correct hole given a per-timestep reward of distance to the hole (see Appendix~\ref{sec:app-sawyer}).
Due to lab access restrictions as a result of COVID-19, we could not evaluate adaptation to new tasks on this platform.
Videos of all experiments can be found on our project website.\footnote{ \url{https://sites.google.com/view/meld-lsm/home}}

\vspace{-0.1in}
\section{Discussion} 
\label{sec:conclusion}

In this paper, we leverage the insight that meta-RL can be cast into the framework of latent state inference. 
This allows us to combine the fast skill-acquisition capabilities of meta-learning with the efficiency of unsupervised latent state models when learning from raw image observations. Based on this principle, we design \MethodNoSpace, a practical algorithm for meta-RL with image observations. \Method outperforms prior methods on simulated locomotion and manipulation tasks, and is efficient enough to perform meta-RL directly from images in the real world. 
However, neither our approach nor prior meta-learning works have shown convincing generalization to wider task distributions of qualitatively distinct manipulation tasks. 
Additionally, due to the difficulty of defining and instrumenting shaped reward functions for such tasks,
it is important that meta-RL algorithms be able to learn from less structured signals and other forms of supervision, such as demonstrations and natural language.
We view these directions as exciting avenues of future work that will broaden the applicability of robotic learning to less structured environments.

\textbf{Acknowledgements}
This research was supported by the NASA Early Stage Innovations program and ARL DCIST CRA W911NF-17-2-0181, with compute support from Google.

\setlength{\bibsep}{0pt plus 0.3ex}
\begin{footnotesize}
\bibliography{MELD}  
\end{footnotesize}
\appendix
\section{MELD Implementation Details}
\label{sec:app-impl}
Here, we expand on Section~\ref{sec:imp} to provide further implementation details of our algorithm MELD. Please also refer to our open-source code:~\url{https://github.com/tonyzhaozh/meld}.

As discussed in Section~\ref{sec:imp}, the latent state model is comprised of latent dynamics distributions, posterior inference distributions, and generative distributions of observations and rewards.
While most timesteps are processed by the same time-invariant dynamics model $p(\state_t | \state_{t-1}, \act_{t-1})$ and posterior inference network $q_{\phi}(\state_t | \state_{t-1}, \obs_t, r_t, \act_{t-1})$, we use separate distributions to model the first time step of a trial: $p(\state_1 | \state_0)$ and $q_{\phi}(\state_1 | \state_0, \obs_1, r_1)$.
Note that in Sections~\ref{sec:exp-button} and ~\ref{sec:exp-real}, we assume trials contain $2$ episodes, while in Section~\ref{sec:exp-sim} trials contain $1$ episode.
Following SLAC, we implement two layers of latent variables (please refer to the SLAC paper~\cite{lee2019slac}, for more details).
The network $q_{\phi}(\state_t | \state_{t-1}, \obs_t, r_t, \act_{t-1})$ encodes image observations and the network $p(\obs_t | \state_t)$ decodes them for the reconstruction loss. 
Both of these networks include the same convolutional architecture (the decoder simply the transpose of the encoder) that consists of five convolutional layers.
The layers have $32$, $64$, $128$, $256$, and $256$ filters and the corresponding filter sizes are $5$, $3$, $3$, $3$, $4$.
For environments in which the robot observes two images (such as scene and wrist camera), we concatenate the images and apply rectangular filters.
All other model networks are fully connected and consist of $2$ hidden layers of $32$ units each.
We use ReLU activations after each layer.

As discussed, we train the actor and critic using the SAC RL algorithm~\cite{haarnoja2018soft} with the belief state as input.
The SAC algorithm maximizes discounted returns as well as policy entropy via policy iteration.
The critic (Q-function) is trained to minimize the soft Bellman error, which takes the entropy of the policy into account in the backup.
We instantiate the actor and critic as fully connected networks with $2$ hidden layers of $256$ units each. 
We follow the implementation of SAC, including the use of $2$ $Q$-networks and the $\tanh$ actor output activation (please see the SAC paper~\cite{haarnoja2018soft} for more details).

Meta-training alternates between collecting data with the current model and policy, and training the model and actor-critic.
Gradients from the actor-critic optimization do not flow into the latent state model.
Before beginning this alternating scheme, we first train the latent state model on $60$ trajectories of data collected with a random policy.
Relevant hyper-parameters for meta-training can be found in Table~\ref{tab:params}.

\begin{table}[h]
\centering
\caption{Meta-training hyper-parameters}
\begin{tabular}{ l l }
\hline
 Parameter & Value \\ 
 \hline
 num. training tasks & $30$ \\
 num. eval tasks & $10$ \\
 actor, critic learning rates & $3\text{e-}4$ \\  
 model learning rate & $1\text{e-}4$ \\
 size of per-task replay buffers $\mathcal{B}_i$ & $1\text{e}5$ \\
 num. tasks collect data & $20$ \\
 num. rollouts per task & $1$ \\
 num. train steps per epoch & $640$ \\
 num. tasks sample for update & $20$ \\
 model batch size & $512$ \\
 actor, critic batch size & $512$ \\
 \hline
 \label{tab:params}
\end{tabular}
\end{table}

\newpage
\section{Simulation Experiment Details and Ablation Study}
\label{sec:app-sim}
In this section we provide further details on the simulated experiments in Section~\ref{sec:exp-sim}. We also perform an ablation study of several design decisions in the MELD algorithm.

\subsection{Success Metrics}
We define a success metric for each environment that correlates with qualitatively solving the task: Cheetah-vel: within $.2$m/s of target velocity, Reacher: within $10$cm of goal, Peg: complete insertion with $5$cm variation possible inside the site, Shelf: mug within $5$cm of goal.
We use these success metrics because task reward is often misleading when averaged across a distribution of tasks; in peg-insertion, for example, the numerical difference between always inserting the peg correctly versus never inserting it can be as low as $0.1$, since the distance between the center of the goal distribution and each goal is quite small and accuracy is required.
In Figure~\ref{fig:exp-comparisons}, we plot this success threshold as a dashed black line.

\subsection{Environment Details}
\begin{figure}[H]
    \centering
    \includegraphics[height=0.2\textwidth]{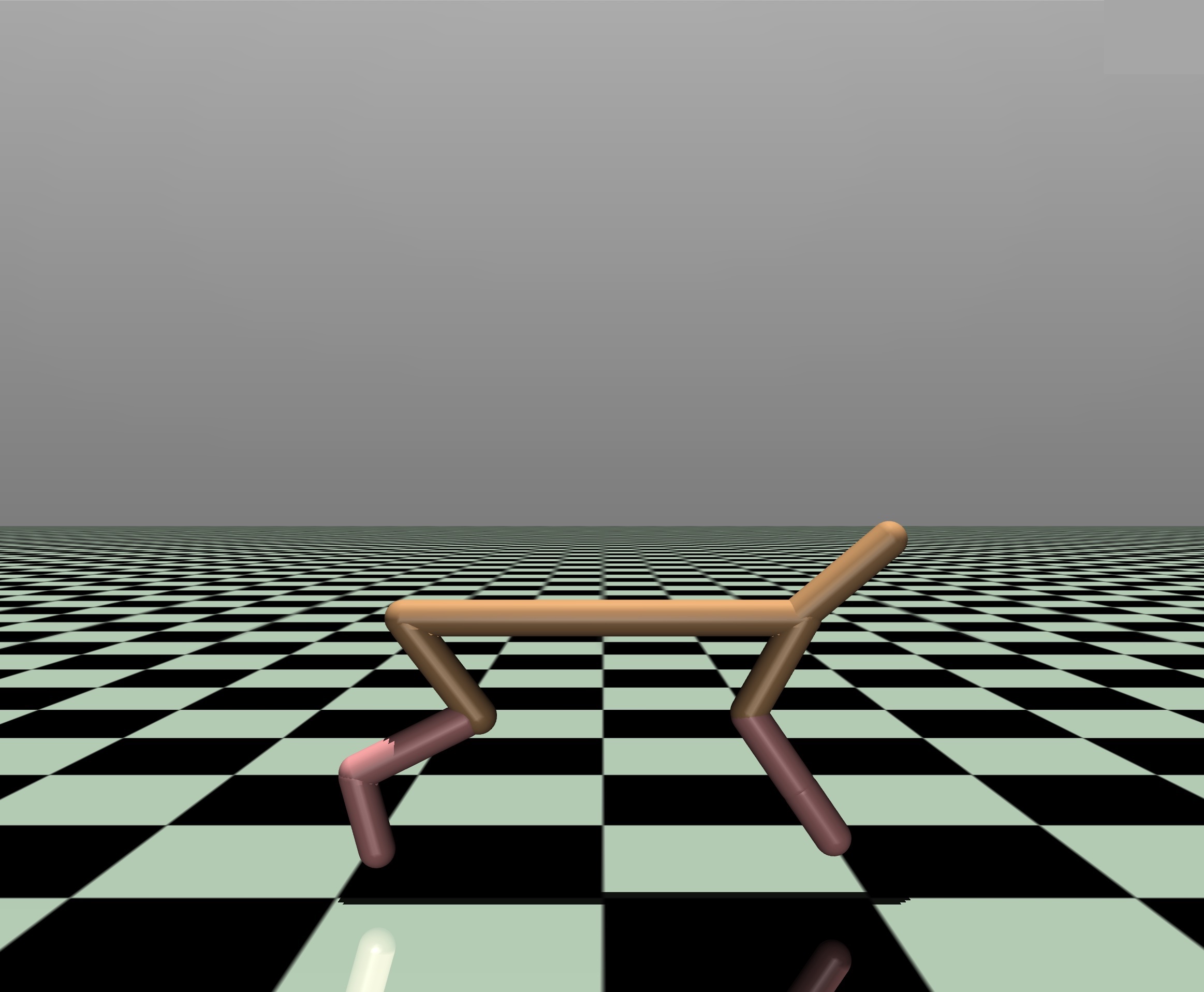}\\
    \vspace{4pt}\includegraphics[height=0.2\textwidth]{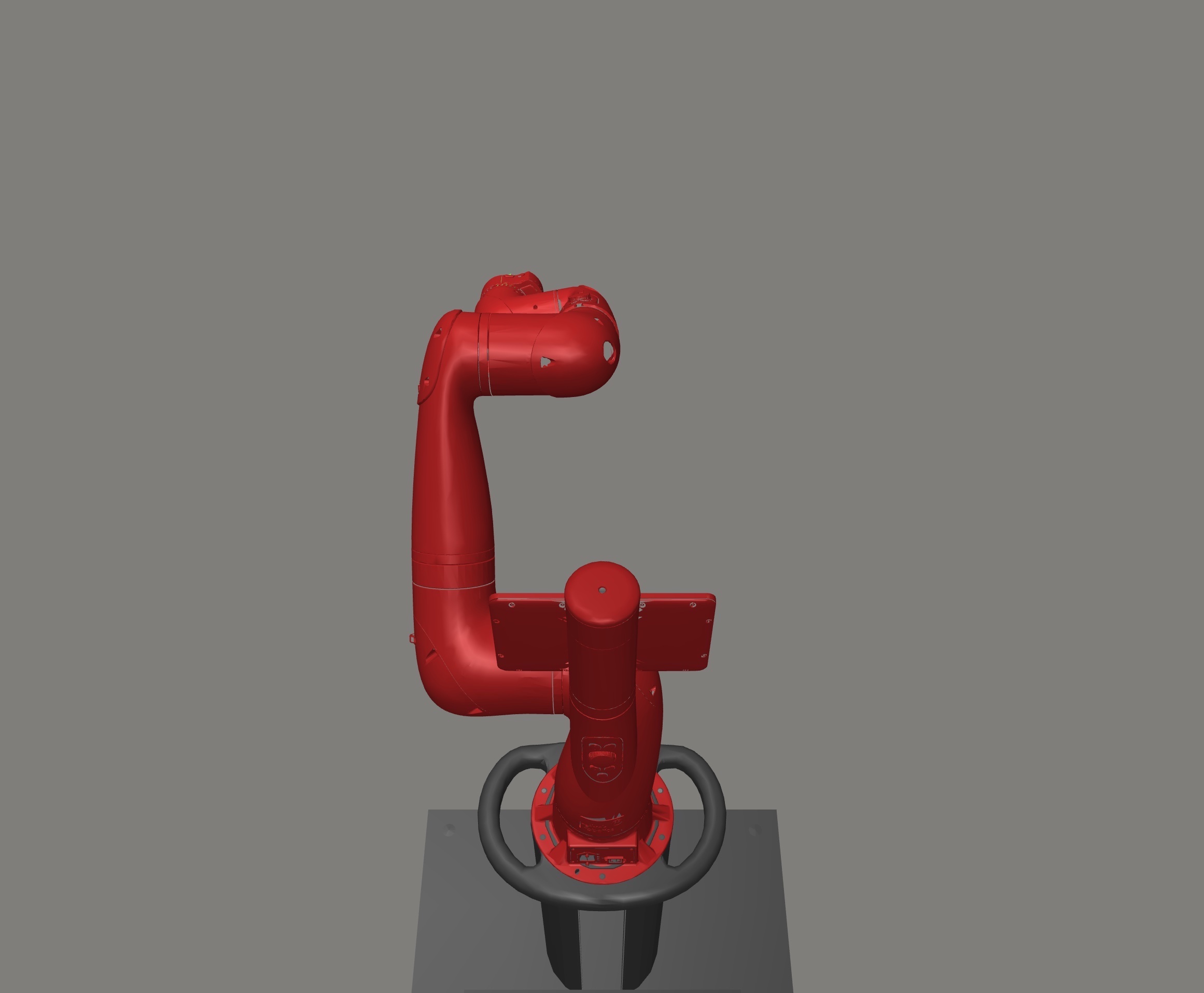}\hspace{-4pt}
    \includegraphics[height=0.2\textwidth]{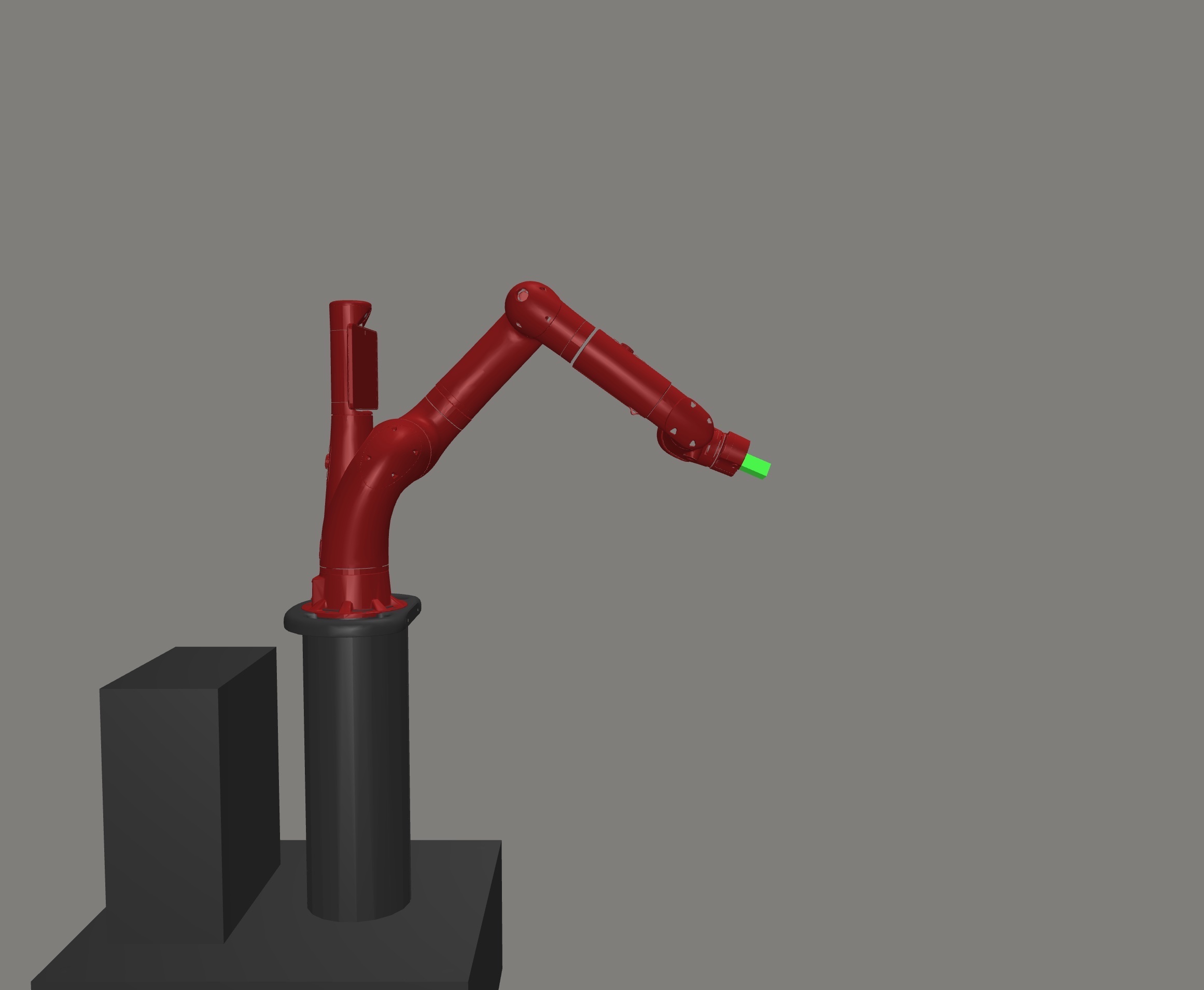}\\
    \vspace{4pt}\includegraphics[height=0.2\textwidth]{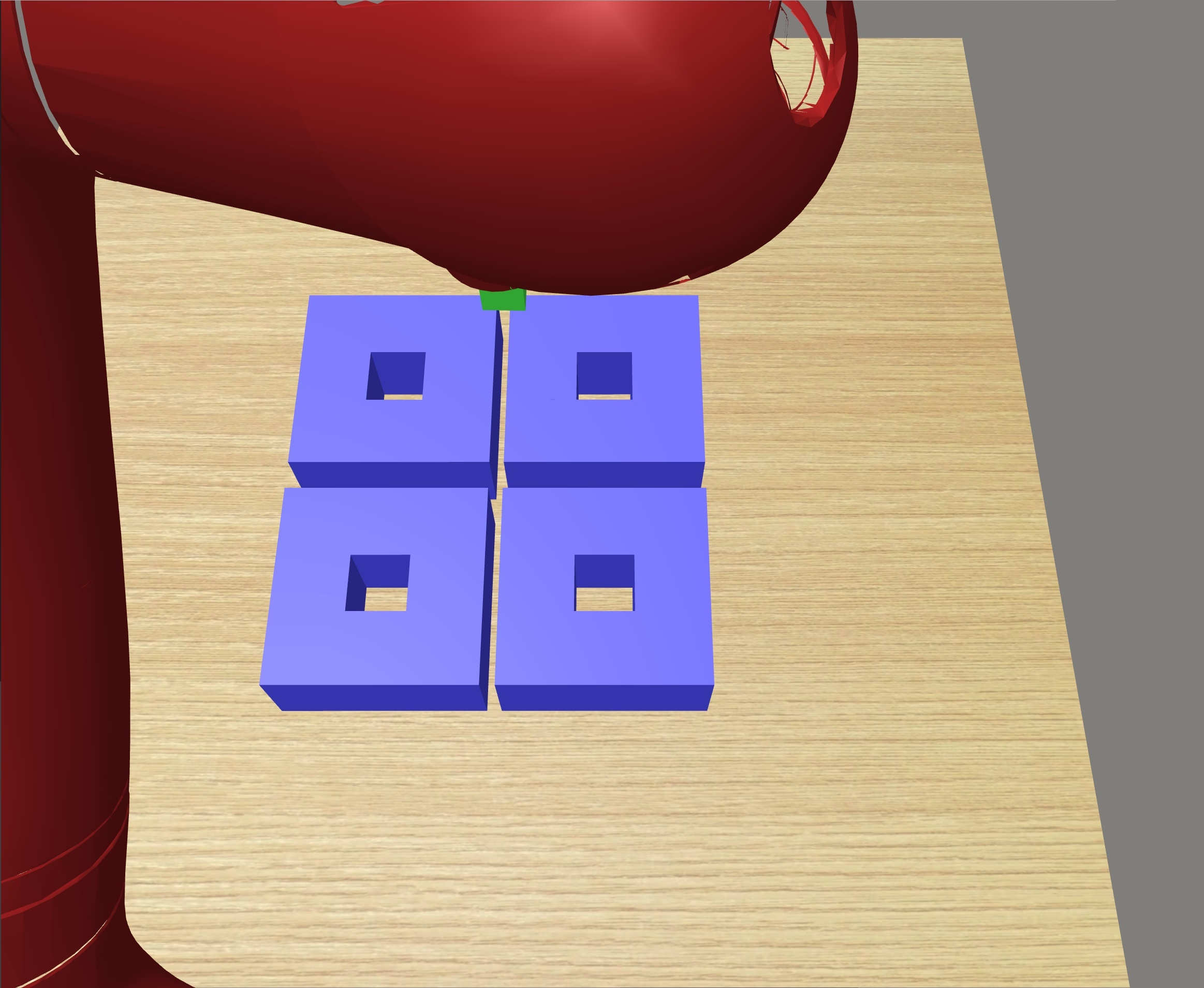}\hspace{-4pt}
    \includegraphics[height=0.2\textwidth]{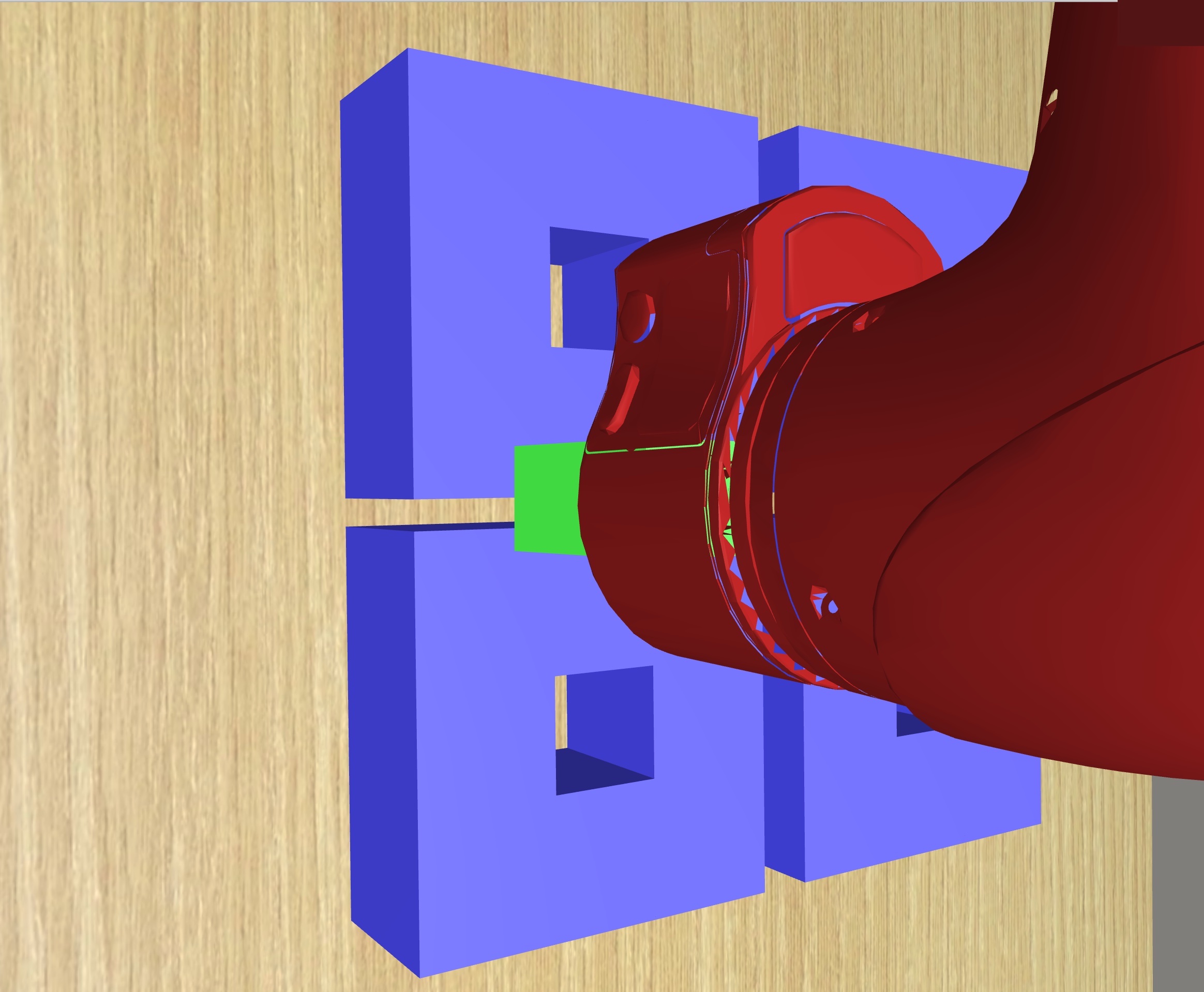}\\
    \vspace{4pt}\includegraphics[height=0.2\textwidth]{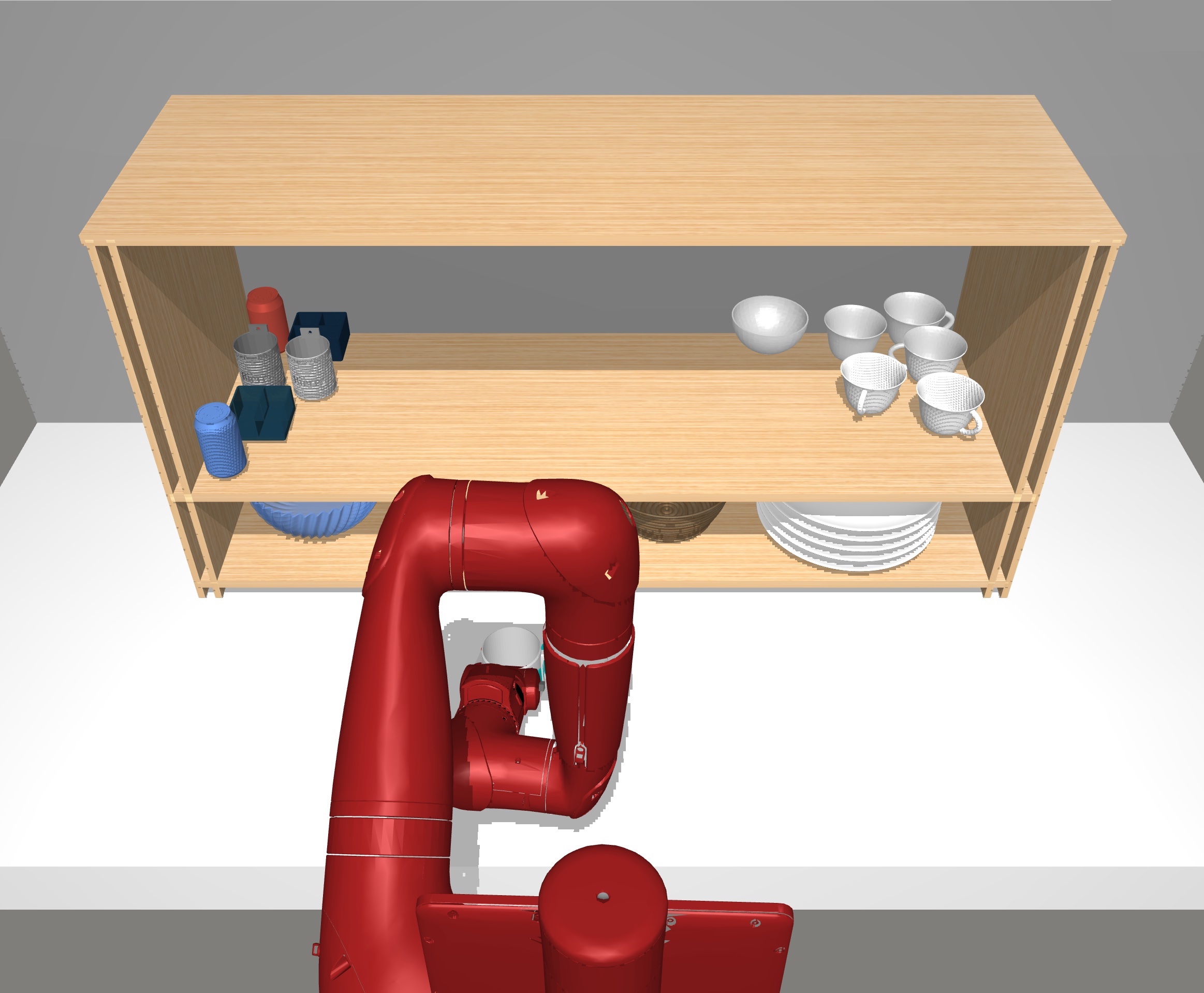}\hspace{-4pt}
    \includegraphics[height=0.2\textwidth]{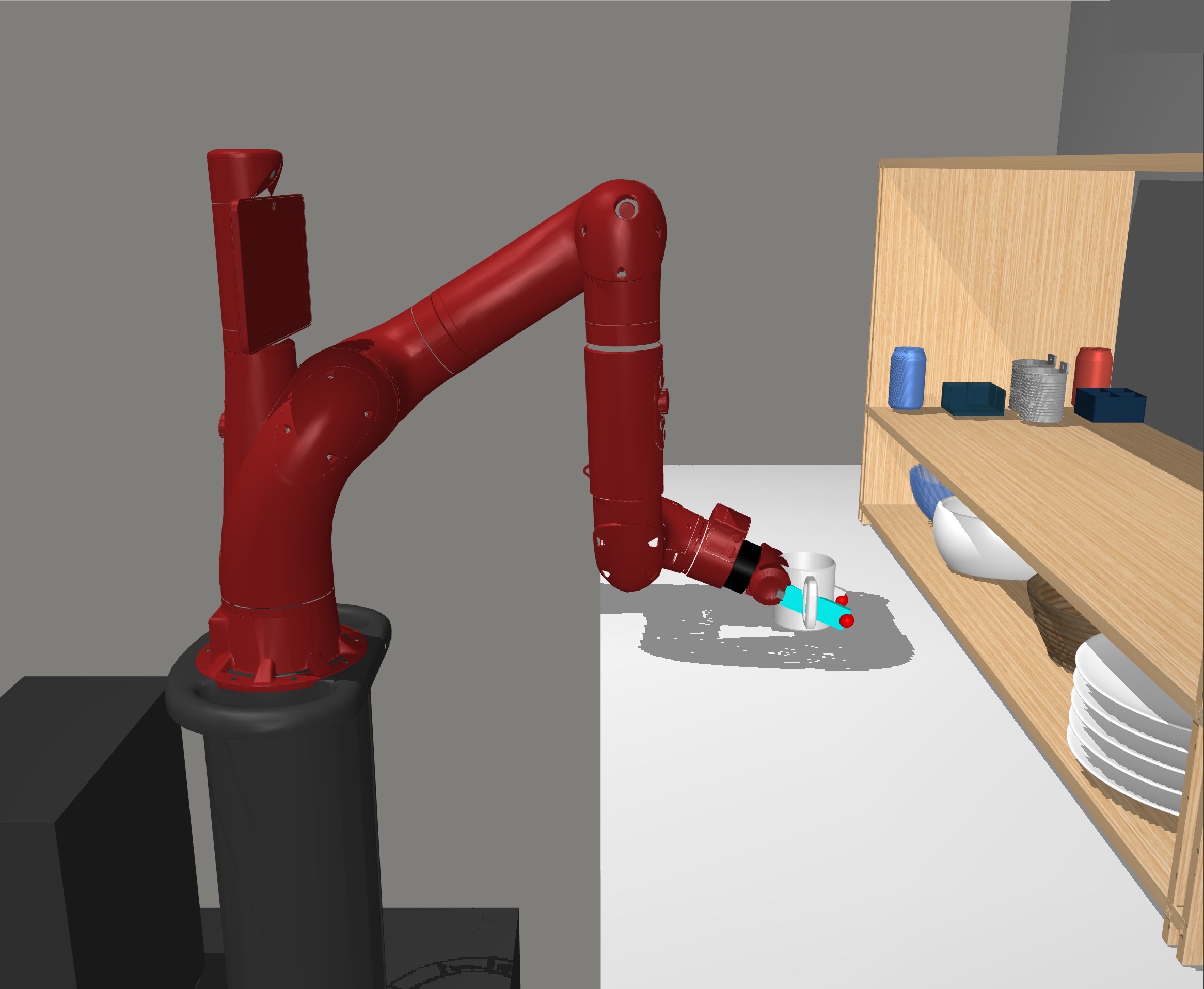}
    \caption{$64$x$64$ and $64$x$128$ image observations, seen as input by MELD for (a) Cheetah-vel, (b) Reacher, (c) Peg-insertion, and (d) Shelf-placing.}
    \label{fig:all-obs}
\end{figure}

In the Cheetah-vel environment, we control the robot by commanding the torques on the robot's $6$ joints.
The reward function consists of the difference between the target velocity $v_\text{target}$ and the current velocity $v_x$ of the cheetah's center of mass, as well a small control cost on the torques sent to the joints:
\begin{align}
r_\text{cheetal-vel} = -|v_x - v_\text{target}| + 0.01||a_t||_2.
\end{align}
The episode length is $50$ time steps, and the observation consists of a single $64$x$64$ pixel image from a tracking camera (as shown in Fig.~\ref{fig:all-obs}a),
which sees a view of the full cheetah.

In all three Sawyer environments, we control the robot by commanding joint delta-positions for all $7$ joints. 
The reward function indicates the difference between the current end-effector pose $x_\text{ee}$ and a goal pose $x_\text{goal}$, as follows:
\begin{align}
r_\text{sawyer-envs} = -(d^2 + \log(d + 1\text{e-}5)), \hspace{8pt} \text{where} \hspace{4pt} d=||x_\text{ee}-x_\text{goal}||_2.
\end{align}
This reward function encourages precision near the goal, which is particularly important for the peg insertion task.
We impose a maximum episode length of $40$ time steps for these environments.
The observations for all three of these environments consist of two images concatenated to form a $64$x$128$ image: one from a fixed scene camera, and one from a wrist-mounted first-person view camera.
These image observations for each environment are shown in Fig.~\ref{fig:all-obs}b-d.
The simulation time step and control frequency for each of these simulated environments is listed in Table~\ref{tab:envs}.

\begin{table}[h]
\centering
\caption{Simulation Environments}
\begin{tabular}{ l l l }
\hline
 Environment & Sim. time step & Control freq. \\ 
 \hline
 Cheetah-vel & $0.01$ & $10$Hz \\
 Reacher & $0.0025$ & $4$Hz \\
 Peg-insert & $0.0025$ & $4$Hz \\
 Shelf-placing & $0.0025$ & $4$Hz \\
 \hline
 \label{tab:envs}
\end{tabular}
\end{table}

\subsection{Ablation Study}
\begin{figure}[H]
\begin{minipage}{0.31\linewidth}
    \includegraphics[width=0.99\textwidth]{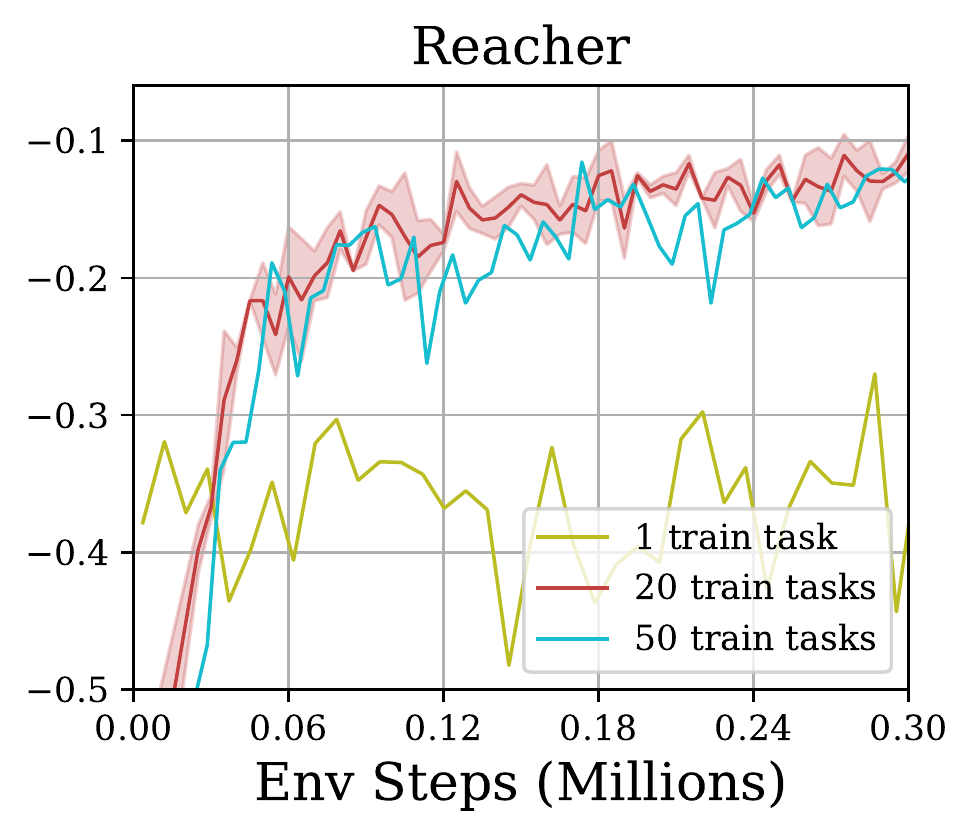}
    \caption{\textbf{Number of meta-training tasks} Generalization to evaluation tasks requires training across set of tasks, but increasing beyond $20$ tasks does not improve.}
    \label{fig:generalization}
\end{minipage}\hfill
\begin{minipage}{0.31\linewidth}
    \vspace{0.15in}
    \includegraphics[width=0.99\textwidth]{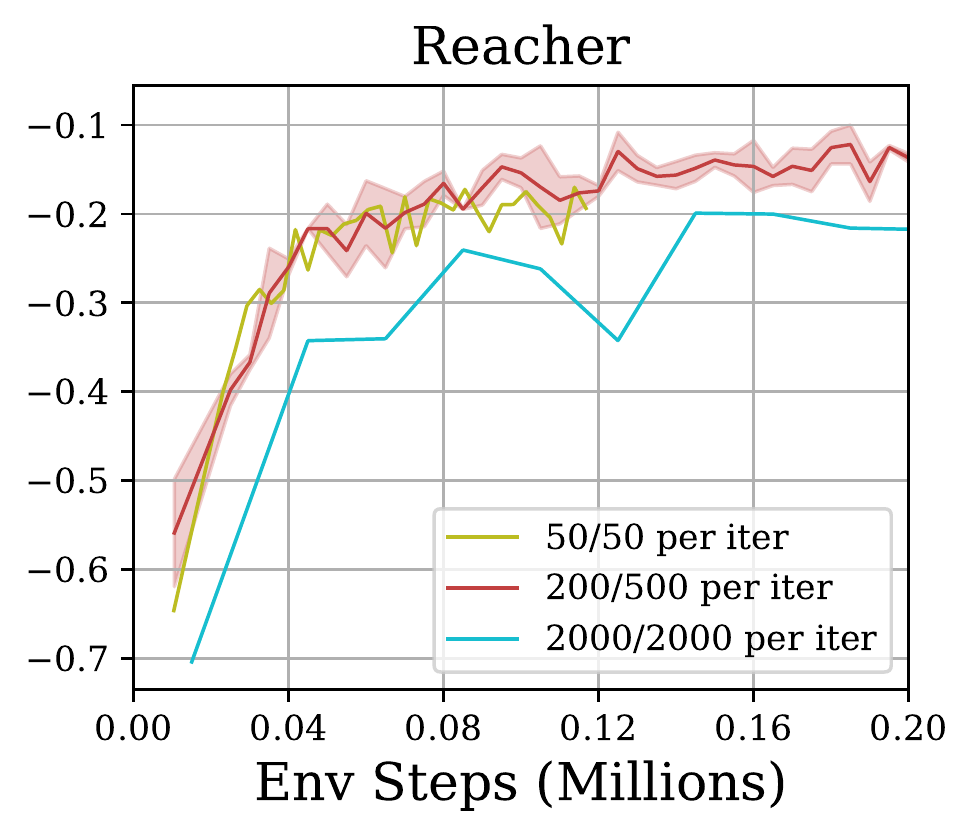}
    \caption{\textbf{Data / gradients ratio} \Method is not very sensitive to scaling the batch size of data collected, along with the amount of actor-critic training, at each iteration.}
    \label{fig:data-collected}
\end{minipage}\hfill
\begin{minipage}{0.31\linewidth}
    \vspace{-0.1in}
    \includegraphics[width=0.99\textwidth]{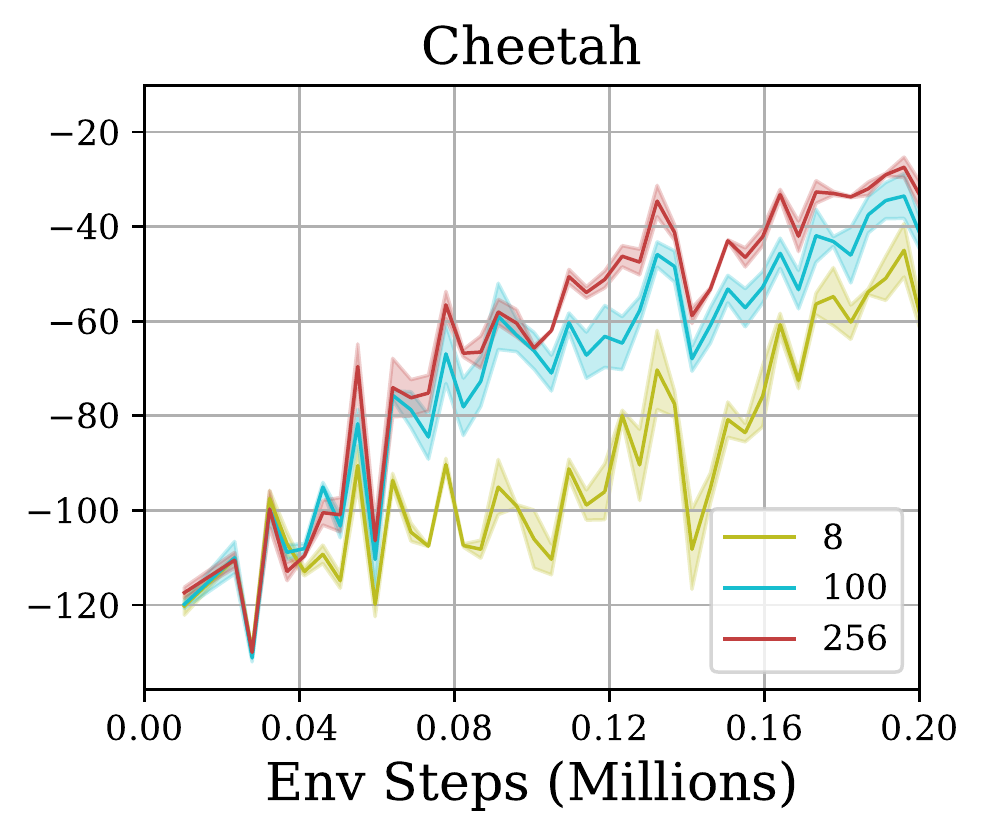}
    \caption{\textbf{Latent variable dimension} Latent state dimension affects task performance, larger is better.
    }
    \label{fig:latent-dim}
\end{minipage}
\end{figure}

In this section we ablate several hyper-parameters of the algorithm to understand which components affect its performance.
First, we examine the effect of the number of meta-training tasks on MELD's performance on test tasks.
In Figure~\ref{fig:generalization} we plot the performance on test tasks for different numbers of train tasks on the Reacher problem.
We find that while generalization fails completely with only a single meta-training task, there are diminishing returns for increasing the number of training tasks beyond $20$.
This result demonstrates promising generalization, since a relatively low number of training tasks is actually needed in order to be able to solve the held-out test tasks. This result also has a practical benefit in that it precludes the need for instrumenting a large task distribution for meta-training in the real world, which can be cumbersome.

Although the reference implementation of SAC calls for each data collection step to be interleaved with a training step, this process of stopping after each environment step in order to perform training is not possible in the real world.
Instead, in MELD, we collect batches of data that consist of full episodes, and we interleave these collection phases with training phases that consist of many gradient steps. Here, we examine the affect of this choice on the performance of the algorithm in simulation.
From the results shown in Figure~\ref{fig:data-collected}, we see that MELD is not too sensitive to scaling between this type of batched off-policy training and the original SAC-style on-policy training,
which is essential for training in the real world.

In Figure~\ref{fig:latent-dim}, we examine the effect of the dimension of the latent variable $\state_t$.
These experiments show that a larger latent dimension performs better; we use dimension $256$ in all our experiments.

\section{Sparse Reward Method and Experiment Details}
\label{sec:app-sparse}
As described in Section~\ref{sec:exp-button}, when reward is given only upon completion of the task, efficient exploration is required to identify a new task within a few trials.
The agent can acquire these exploration strategies during meta-training by learning strategies tailored to the task distribution.
For example, in the button-pressing problem presented in Section~\ref{sec:exp-button}, a learned exploration strategy might try pushing each button in succession, but would not try to e.g., pick up the control panel.

While in principle these behaviors can be acquired by MELD as described in Section~\ref{sec:imp}, in practice performing RL with sparse rewards at meta-training time presents a significant exploration challenge.
In effect, to learn useful exploration strategies for meta-test time, the agent must first explore effectively during meta-training.
Because  RL  with sparse rewards is very inefficient, we follow prior work~\cite{gupta2018meta, rakelly2019efficient} and assume access to a shaped reward function during meta-training to help learn these strategies.
This setup corresponds to a setting where meta-training is performed in a laboratory with access to instrumentation to calculate the shaped reward, while meta-testing occurs outside the lab where only sparse rewards are available.

To make use of the shaped reward during meta-training time, we follow a two-stage procedure. 
First, we perform meta-training using the shaped reward as prescribed in Section~\ref{sec:imp}. 
We then add data collected by this agent to the replay buffer of a second agent, which is trained with a small modification made to the model training loss from Equation~\ref{eq:elbo}.
Here, the latent state model takes the \textbf{sparse} reward signal as input (to match our desired meta-testing setup), but we still use the shaped reward for the reconstruction target. We denote the shaped reward as $\tilde{r}$ and the sparse reward as $r$, and highlight the difference from Equation~\ref{eq:elbo} in blue.
\begin{multline}
    \mathcal{L}_{model}(\obs_{1:T}, r_{1:T}, \textcolor{blue}{\tilde{r}_{1:T}}, \act_{1:T-1}) =  \mathop{\E}_{\state_{1:T} \sim q_\phi} \sum_{t=1}^{T} \log p_\phi(\obs_{t} | \state_{t}) + \log p_\phi(\textcolor{blue}{\tilde{r}_{t}} | \state_{t}) \\ -  D_\text{KL}(q_\phi(\state_1 | \obs_1, r_1) \| p(\state_1))  - \sum_{t=2}^{T} D_\text{KL}(q_\phi(\state_{t} | \obs_{t}, r_{t}, \state_{t-1}, \act_{t-1}) \| p_\phi(\state_{t} | \state_{t-1}, \act_{t-1})).
\vspace{-0.3in}
\end{multline}
Additionally, we use the shaped reward to train the critic, since this is also not required at meta-test time.

Note that MELD does \emph{not} simply learn to copy the trajectories from the shaped-reward training that were used to warmstart the sparse-reward training. The former (``expert'') trajectories move from the starting position directly to the correct button, since the shaped reward contains information to almost immediately identify the correct task. The MELD trajectories that result from receiving only sparse rewards as input, however, demonstrate systematic exploration of visiting different buttons in order to determine the correct one (see Figure~\ref{fig:exp-button}). Finally, note that we use this same approach for the real-world WidowX experiments in Section~\ref{sec:exp-real}.

\section{Importance of Performing Inference at Every Timestep}
\label{sec:app-time-varying}

In this section, we discuss the benefit of the time-varying nature of MELD's latent belief. We emphasize that this design decision is useful in the standard meta-RL setting, as well as in other realistic settings of the underlying task itself changing within an episode. 

\subsection{Fast Adaptation}

We first present a didactic image-based 2D navigation problem to illustrate how \Method can learn extended exploration strategies to adapt to a new task, similar to the results shown in Figure~\ref{fig:exp-button}. Here, the task distribution consists of goals located along a semi-circle around the start state. The agent receives inputs in the form of $64$x$64$ image observations and rewards that are non-zero only upon reaching the correct goal. 
We use the same approach to using shaped reward for meta-training as described in Appendix~\ref{sec:app-sparse}.
\begin{wrapfigure}{h}{0.5\textwidth}
\centering
\includegraphics[height=0.135\textheight]{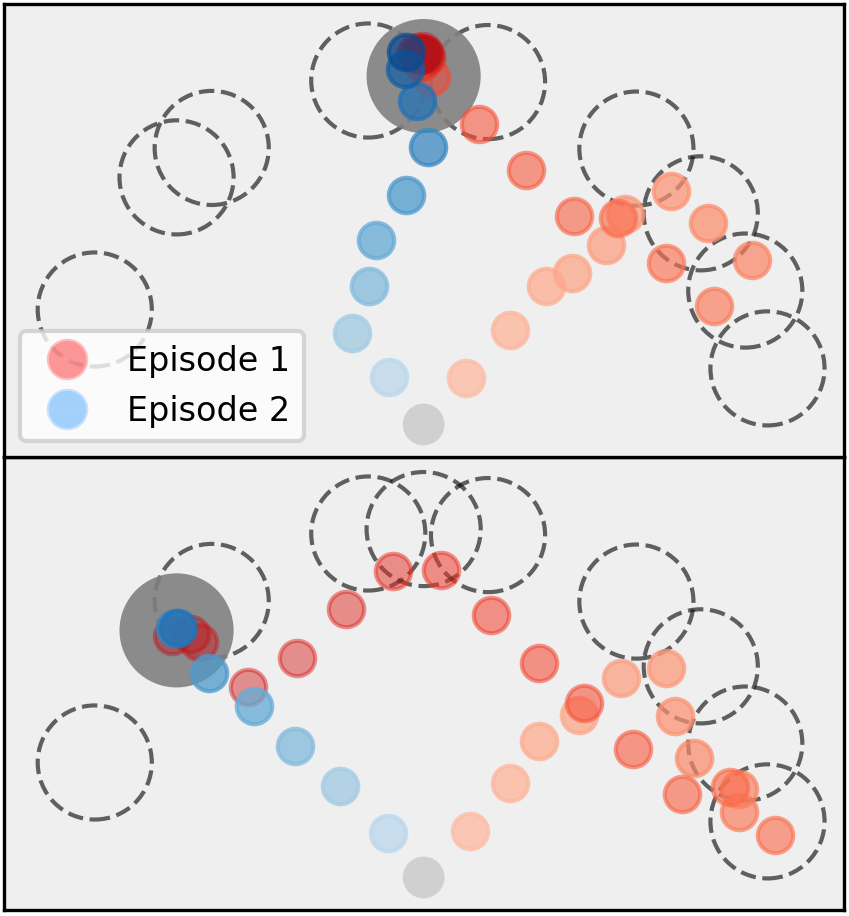}
\includegraphics[height=0.135\textheight]{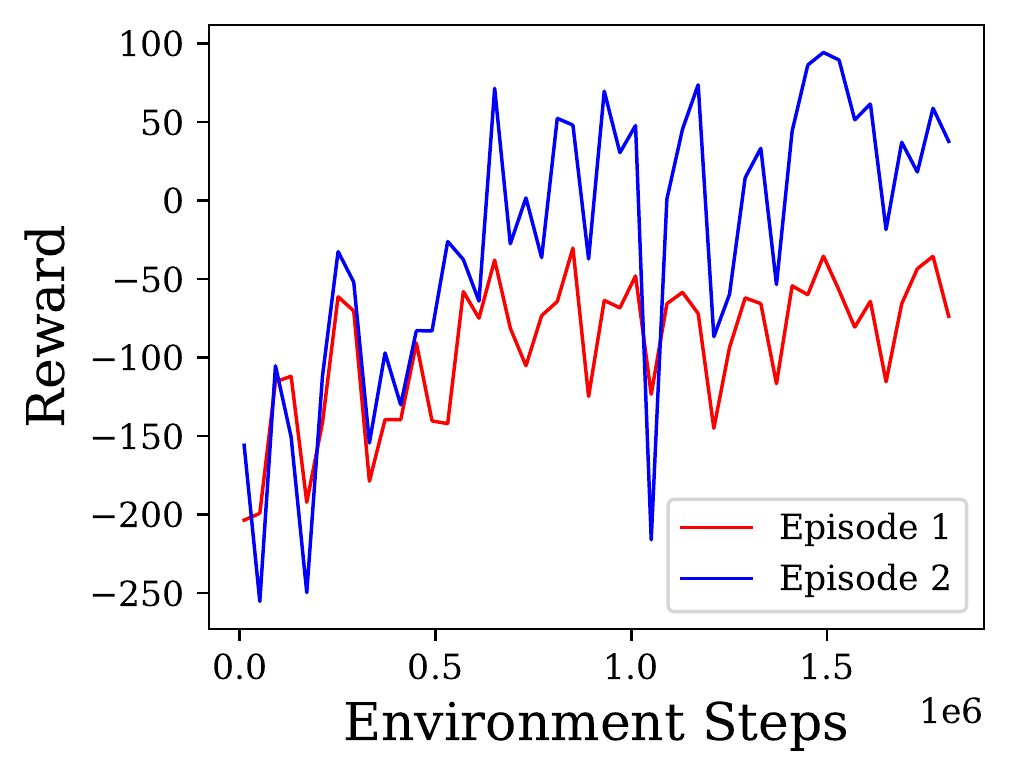}
  \captionof{figure}{
  Image-based 2D navigation with reward given after finding the right goal: (Left) trajectory traces of meta-learned exploration finding goal in Episode-1 (red) and going directly there in Episode-2 (blue). (Right) higher rewards in episode-2 show preservation of information across episodes.
  }\vspace{-0.1in}
  \label{fig:exp-pm}
\end{wrapfigure}

As shown in Figure~\ref{fig:exp-pm}a, the agent learns an efficient exploration strategy of traversing the semi-circle goal region until the goal is found.
By updating the posterior belief at each step, \Method is able to find the goal within $10-20$ steps, instead of multiple episodes as required by methods that explore via posterior sampling~\cite{rakelly2019efficient, gupta2018meta} that hold the task variable constant across each episode. Furthermore, note that once the goal is found, \Method can navigate directly to it (Figure~\ref{fig:exp-pm}b) in the next episode. 

This experiment demonstrates that even in standard meta-RL setting where the underlying task remains constant throughout an episode, updating task information at each timestep can enable faster adaptation. We argue that this behavior has safety benefits in the real world, since the agent need not complete full episodes of potentially hazardous exploration before incorporating task information.

\subsection{Adapting to Task Changes within Episodes}
\begin{wrapfigure}{h}{0.5\textwidth}
\centering\vspace{-0.2in}
  \includegraphics[height=0.15\textheight]{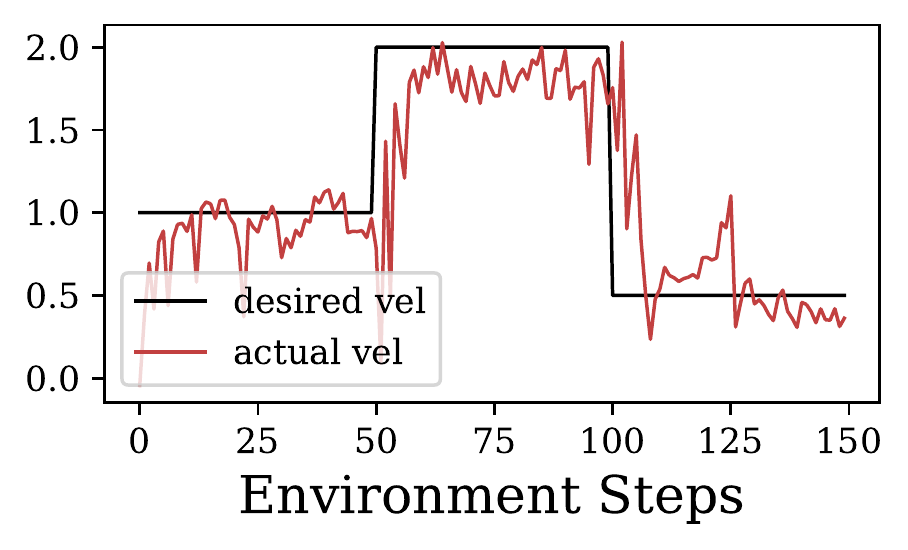}\vspace{-0.1in}
  \caption{\Method tracking changing velocity targets for Cheetah-vel. 
  }
  \label{fig:cheetah-vel-changes}
\end{wrapfigure}
The experiments in the main paper as well as the section above consider the standard meta-RL paradigm, where the agent adapts to one test task at a time.
However, many realistic scenarios consist of a sequence of tasks.
For example, consider a robot moving a mug filled with liquid; if some of the liquid spills, the robot must adapt to the new dynamics of the lighter mug to finish the job.
Because MELD updates the belief over the hidden variables at each time step, it can be directly applied to this setting without modification.
We evaluate MELD in the Cheetah-vel environment on a sequence of $3$ different target velocities within a single episode and observe in Figure~\ref{fig:cheetah-vel-changes} that \Method adapts to track each velocity within a few time steps.

\section{Task Reset Mechanism for WidowX Experiments}
\label{sec:app-widowx}
Since meta-training requires training across a distribution of tasks, we build an automatic task reset mechanism for the real-world experiments with the WidowX robot performing ethernet cable insertion.
This mechanism controls the translational and rotational displacement of the network switch. The network switch(A) is mounted to a $3$D printed housing(B) with gear attached. We control the rotation of the housing through motor $1$. This setup is then mounted on top of a linear rail(C) and motor $2$ controls its translational displacement through a timing pulley. In our experiments, the training task distribution consisted of $20$ different tasks, where each task was randomly assigned from a rotational range of $16$ degrees and a translational range of $2$cm. 

\begin{figure}[h]
  \centering
    \includegraphics[height=0.25\textheight]{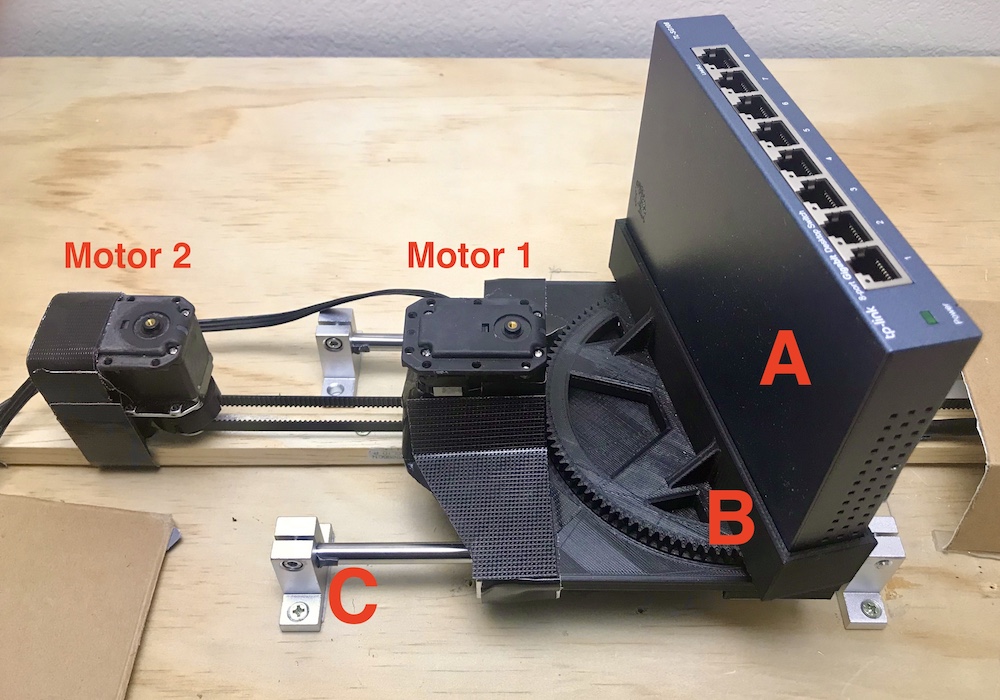}
  \caption{\textbf{Automatic task reset mechanism} The network switch is rotated and translated by a series of motors in order to generate different tasks for meta-learning. This allows our meta-learning process to be entirely automated, without needing human intervention to reset either the robot or the task at the beginning of each rollout.
  }
  \label{fig:hardware-reset}
  \vspace{-8pt}
\end{figure}

\vspace{-.03in}
\section{Sawyer Multi-task Peg Insertion}
\label{sec:app-sawyer}
\begin{wrapfigure}{h}{0.5\textwidth}
  \centering
  \includegraphics[height=0.2\textheight]{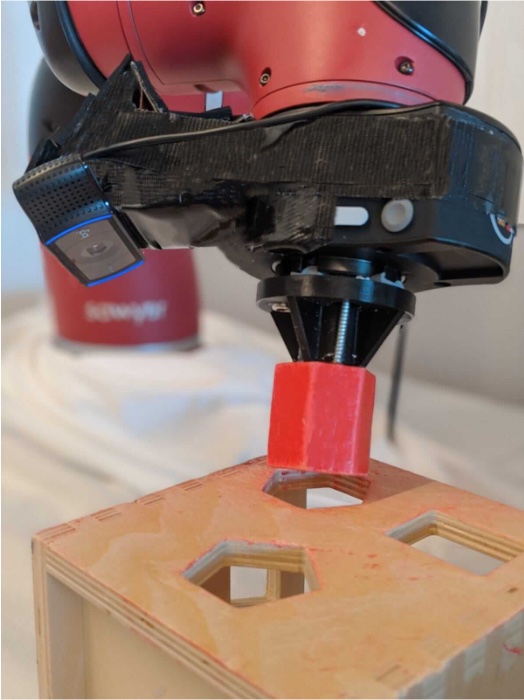}
  \includegraphics[height=0.2\textheight]{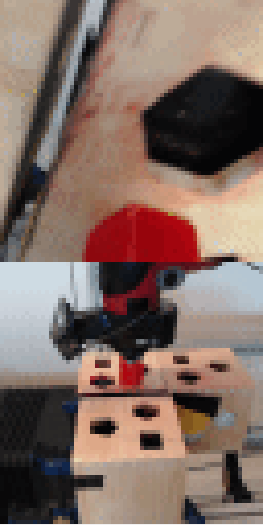}
  \caption{\textbf{(left)} Performing precise peg insertion in the real world with a $7$-DoF Sawyer robot. \textbf{(right)} $64$x$128$ image observations seen as input by the Sawyer.}
  \label{fig:sawyer-task}
\end{wrapfigure}

\begin{figure}[H]
  \centering 
    \includegraphics[height=0.20\textheight]{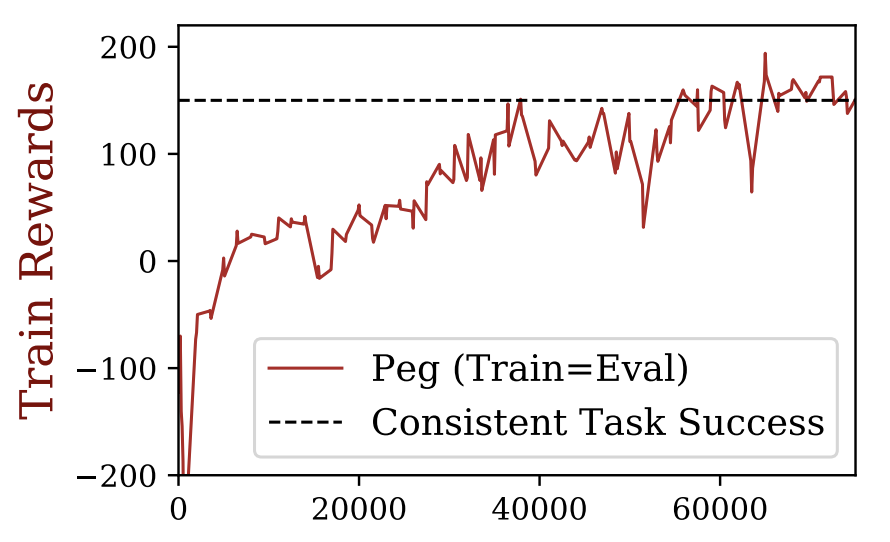}
  \caption{Rewards on train tasks during meta-training for Sawyer peg insertion.
  }
  \label{fig:sawyer-plot}
\end{figure}

In this experiment, we demonstrate that \Method can reason jointly about state and task information to perform real-world peg insertion with a $7$-DoF Sawyer robot (Figure~\ref{fig:sawyer-task}, left). On the Sawyer robot, we learn precise peg insertion where the task distribution consists of three tasks, each corresponding to a different target box. 
Note that the goal is not provided to the agent, but must be inferred from its history of observations and rewards. 
The reward function is the sum of the L$2$-norms of translational and rotational distances between the pose of the object in the end-effector and a goal pose.
The agent's observations are concatenated images from two webcams (Figure~\ref{fig:sawyer-task}, right): one fixed view and one first-person view from a wrist-mounted camera. The robot succeeds on all three tasks after training on $4$ hours of data ($60,000$ samples at $4$Hz), as shown in Figure~\ref{fig:sawyer-plot}.
Videos of the experiment may be found on the project website.

\end{document}